\documentclass[runningheads]{llncs}

\usepackage{eccv}

\usepackage{eccvabbrv}

\usepackage{graphicx}
\usepackage{booktabs}

\usepackage[accsupp]{axessibility}  % Improves PDF readability for those with disabilities.

\usepackage[pagebackref,breaklinks,colorlinks,citecolor=eccvblue]{hyperref}
\usepackage{orcidlink}

\usepackage{algorithm}
\usepackage{algorithmic}
\usepackage{makecell}

\usepackage[most]{tcolorbox}
\definecolor{MorandiBlue}{RGB}{235, 241, 245}
\definecolor{MorandiBlueFrame}{RGB}{163, 184, 196}

\tcbuselibrary{breakable}

\begin{document}

% ---------------------------------------------------------------
% % TODO REVIEW: Replace with your title
% \title{\textsc{EruDiff}: Refactoring Knowledge in Diffusion Models for Advanced Text-to-Image Synthesis} 

% % TODO REVIEW: If the paper title is too long for the running head, you can set
% % an abbreviated paper title here. If not, comment out.
% \titlerunning{\textsc{EruDiff}: Refactoring Knowledge in Diffusion Models for Advanced T2I}

% % TODO FINAL: Replace with your author list. 
% % Include the authors' OCRID for the camera-ready version, if at all possible.
% Second Author\inst{2,3}\orcidlink{1111-2222-3333-4444} \and
% Third Author\inst{3}\orcidlink{2222--3333-4444-5555}}

% % TODO FINAL: Replace with an abbreviated list of authors.

% % TODO FINAL: Replace with your institution list.
% Springer Heidelberg, Tiergartenstr.~17, 69121 Heidelberg, Germany

\title{\textsc{EruDiff}: Refactoring Knowledge in Diffusion Models for Advanced Text-to-Image Synthesis} 

\titlerunning{\textsc{EruDiff}: Refactoring Knowledge in Diffusion Models for Advanced T2I}

\author{Xiefan Guo\inst{1,2}\quad 
Xinzhu Ma\inst{1,2}\quad 
Haoxiang Ma\inst{2}\quad 
Zihao Zhou\inst{1}\quad 
Di Huang\inst{1}\thanks{Corresponding author.}
}

\authorrunning{X.~Guo \emph{et al.}}

\institute{\textsuperscript{1}Beihang University\quad 
\textsuperscript{2}Shanghai AI Laboratory
}

\maketitle

\begin{abstract}
  Text-to-image diffusion models have achieved remarkable fidelity in synthesizing images from explicit text prompts, yet exhibit a critical deficiency in processing implicit prompts that require deep-level world knowledge, ranging from natural sciences to cultural commonsense, resulting in counter-factual synthesis. This paper traces the root of this limitation to a fundamental dislocation of the underlying knowledge structures, manifesting as a chaotic organization of implicit prompts compared to their explicit counterparts. In this paper, we propose \textsc{EruDiff}, which aims to refactor the knowledge within diffusion models. Specifically, we develop the Diffusion Knowledge Distribution Matching (DK-DM) to register the knowledge distribution of intractable implicit prompts with that of well-defined explicit anchors. Furthermore, to rectify the inherent biases in explicit prompt rendering, we employ the Negative-Only Reinforcement Learning (NO-RL) strategy for fine-grained correction. Rigorous empirical evaluations demonstrate that our method significantly enhances the performance of leading diffusion models, including FLUX and Qwen-Image, across both the scientific knowledge benchmark (\emph{i.e.}, \textsc{Science}-T2I) and the world knowledge benchmark (\emph{i.e.}, WISE), underscoring the effectiveness and generalizability. Our code is available at \url{https://github.com/xiefan-guo/erudiff}.
  \keywords{Diffusion models \and Text-to-image synthesis \and World knowledge informed image synthesis}
\end{abstract}

\section{Introduction}
\label{sec:introduction}

\begin{figure}[tb]
\centering
\setlength{\belowcaptionskip}{-0.3cm}
\includegraphics[width=0.99\linewidth]{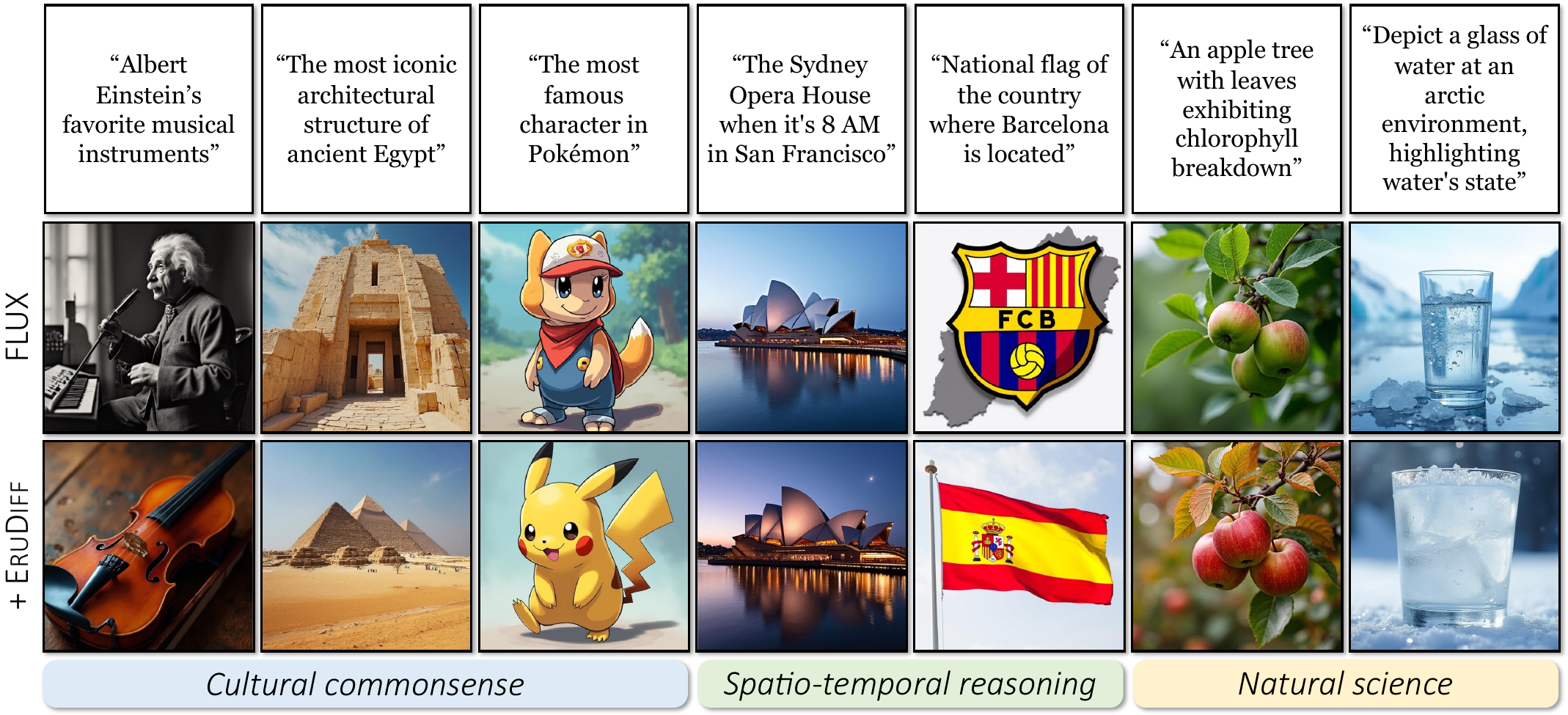}
\caption{\textbf{Visual results on WISE.} \textsc{EruDiff} exhibits a comprehensive integration of multifaceted world knowledge, spanning natural science to cultural commonsense.}
\label{fig:5_qualitative_comparison_flux_wise}
\end{figure}

Text-to-image diffusion models \cite{rombach2022high,betker2023improving,esser2024scaling,chen2024pixart,podell2024sdxl,li2024hunyuan,flux2024,wu2025qwen} have demonstrated substantial efficacy in synthesizing high-fidelity, visually-realistic, and diverse images conditioned on text prompts. Recent breakthroughs, characterized by complex compositional generation \cite{chefer2023attend,liu2022compositional,feng2023training,huang2023t2i,xiao2025fastcomposer,rassin2023linguistic,meral2024conform,guo2024initno,guo2026ctcal}, enhanced aesthetic quality \cite{wu2023human,kirstain2023pick,schuhmann2022laoin,fan2024reinforcement,xu2024imagereward,eyring2024reno,clark2024directly,deng2024prdp,guo2025shortft,karthik2025scalable,liang2025aesthetic}, ultra-high-resolution synthesis \cite{chen2024pixart-sigma,zhang2025diffusion,xie2024sana,zuo20254kagent,he2023scalecrafter,qiu2025freescale}, and efficient architectures and inference acceleration techniques \cite{liu2022flow,geng2025mean,frans2024one,sauer2024adversarial,yin2024one,yin2024improved,salimans2022progressive,song2023consistency,luo2023latent}, have further pushed the boundaries of the field. Despite these improvements, current models exhibit fundamental limitations in scenarios necessitating structured reasoning grounded in world knowledge \cite{niu2025wise,li2025science,sun2025t2i,meng2024phybench,fu2024commonsense}. Their proficiency remains largely confined to the precise rendering of explicit prompts involving superficial attributes, such as color, texture, and shape. Conversely, they demonstrate a restricted capacity to process implicit prompts that encompass a broad spectrum of world knowledge, ranging from natural sciences to cultural commonsense. This deficiency frequently results in counter-factual synthesis, as illustrated in Fig.~\ref{fig:5_qualitative_comparison_flux_wise}.

Current mainstream research paradigms lean towards deep mining the intrinsic world knowledge and reasoning capabilities of Large Language Models (LLMs) to address the cognitive bottlenecks inherent in diffusion models when processing implicit prompts. A representative technical trajectory \cite{wu2024self,yang2024idea2img,manas2024improving,wang2024genartist,qin2024diffusiongpt,wang2025promptenhancer,kou2026think} involves leveraging LLMs to rewrite highly abstract, implicit instructions into granular, explicit descriptions, thereby providing concrete semantic guidance for text-to-image diffusion models to achieve precise synthesis of complex latent meanings. Another frontier seeks to establish unified multimodal understanding and generation frameworks \cite{deng2025emerging,chen2025janus,cao2025hunyuanimage,wang2024emu3,hurst2024gpt,jiang2025t2i,chen2025blip3,xie2024show,ge2024seed,zhou2024transfusion,pan2025transfer}, which involve the cross-modal transfer of deep reasoning capabilities from the textual domain, harnessing inferential power from linguistic modalities to enhance the parsing precision and alignment of visual generative models regarding implicit intentions. Despite these advancements, significant unexplored potential remains within the diffusion models.

This paper posits that a fundamental cause of this performance gap lies in the inherent disparity between training data distributions: while comprehensive world knowledge is predominantly distilled within text-only corpora, text-image pairs are frequently restricted to superficial descriptions of visual appearances. Consequently, we pioneer a novel paradigm that leverages text-only corpora to fine-tune diffusion models, facilitating the internalization of world knowledge.

In this paper, we propose \textsc{EruDiff} ({Erudite Diffusion}) to facilitate a deep refactoring of the inherent knowledge systems within pre-trained diffusion models. This is achieved by leveraging structured text corpora rich in world knowledge, organized into pairs of \emph{implicit prompts} and \emph{explicit prompts}. Specifically, we develop the Diffusion Knowledge Distribution Matching (DK-DM), which achieves precise alignment of knowledge distributions across two semantic spaces by registering high-abstraction implicit prompts to well-defined explicit anchors. To mitigate catastrophic knowledge forgetting, we introduce an anti-forgetting knowledge consolidation mechanism, complemented by a timestep-aware curriculum learning strategy to effectively accelerate convergence. Furthermore, addressing the inherent representation biases and inaccurate visual rendering associated with explicit prompts, we employ the Negative-Only Reinforcement Learning (NO-RL) for fine-grained correction, leading to improved results.

To address the scarcity of structured training resources despite the abundance of world-knowledge evaluation benchmarks \cite{niu2025wise,sun2025t2i,meng2024phybench,fu2024commonsense}, we introduce the Knowledge-10K dataset to serve as the training data support for our method, which will be publicly released to facilitate community research. Extensive empirical evaluations demonstrate that the proposed method significantly enhances the performance of leading diffusion models, including FLUX and Qwen-Image, on the scientific knowledge benchmark (\emph{i.e.}, \textsc{Science}-T2I) and the world knowledge benchmark (\emph{i.e.}, WISE), which validates the superior efficacy and generalization capabilities of our approach.

Our research demonstrates that classical text-to-image diffusion models possess substantial latent cognitive capacity. Systematically refactoring the intrinsic knowledge frameworks unlocks high-dimensional world-knowledge comprehension and superior visual rendering capabilities. This study provides novel insights into the development of next-generation unified multimodal architectures that seamlessly integrate understanding, reasoning, and visual generation.

\section{Related Work}
\label{sec:related_work}

\noindent \textbf{Text-to-image diffusion models.} Text-to-image synthesis aims to generate photorealistic images that align precisely with text prompts. Recently, Diffusion Models (DMs) \cite{ho2020denoising,dhariwal2021diffusion} have become a dominant paradigm in generative modeling, showing exceptional performance in diverse applications \cite{rombach2022high,guo2024i4vgen,zhang2023adding,seedance2025seedance,kong2024hunyuanvideo,huang2023make,wan2025wan,kong2020diffwave,pooledreamfusion}, most notably in text-to-image synthesis. Early landmarks such as GLIDE \cite{nichol2022glide}, DALL-E 2 \cite{ramesh2022hierarchical}, and Imagen \cite{saharia2022photorealistic} explored text-guided refinement, CLIP-based embedding spaces, and cascaded architectures, respectively. Notably, Stable Diffusion (SD) \cite{rombach2022high} significantly enhanced computational efficiency through latent-space diffusion. To further push the boundaries of generation quality, recent research has pivoted toward architectural innovations, including the construction of flow-based methods, the development of Diffusion Transformers (DiT), the emergence of Multi-Modal Diffusion Transformer (MM-DiT) and the integration of large language models (LLMs). Representative state-of-the-art models include PixArt-$\alpha$ \cite{chen2024pixart}, Stable Diffusion 3 \cite{esser2024scaling}, FLUX.1 \cite{flux2024}, HiDream-I1 \cite{cai2025hidream}, Qwen-Image \cite{wu2025qwen}, HunyuanImage \cite{cao2025hunyuanimage} and Seedream \cite{seedream2025seedream}.

Nevertheless, these models encounter fundamental limitations in scenarios requiring structured reasoning informed by world knowledge. Their efficacy is largely restricted to the high-fidelity rendering of explicit prompts concerning surface attributes. In contrast, they exhibit a diminished capacity for addressing implicit prompts rooted in extensive world knowledge, frequently yielding counter-factual synthesis.

\noindent \textbf{Prompt rewriting for diffusion models.} Prompt rewriting aims to augment initial instructions by leveraging the semantic modeling capabilities of Large Language Models (LLMs) before they are passed to a frozen text-to-image (T2I) model, thereby enhancing the quality of visual synthesis. Recent studies \cite{wu2024self,yang2024idea2img,wang2024genartist,qin2024diffusiongpt} utilize LLMs to iteratively critique and refine prompts within a closed-loop system, while others \cite{hao2023optimizing,wu2025reprompt,wang2025promptenhancer,kou2026think} introduced visual feedback signals as rewards to specifically fine-tune LLM-based prompt rewriters. Despite these advancements, the potential of diffusion models remains substantially under-explored. Our research demonstrates that classical text-to-image diffusion models harbor significant latent cognitive capacities that can be effectively unlocked through meticulously designed knowledge refactoring.

\noindent \textbf{RL for diffusion models.} As a direct intervention strategy to enhance the performance of pre-trained diffusion models, Reinforcement Learning (RL) approaches \cite{lee2023aligning,dong2023raft,black2023training,fan2024reinforcement,wallace2024diffusion,yang2024using,li2024aligning,liu2025flow,xue2025dancegrpo,li2025mixgrpo,he2025tempflow,clark2024directly,guo2025shortft,zheng2025diffusionnft} have demonstrated significant efficacy in improving visual quality and text-image alignment. However, its generalizability remains constrained by the scarcity of high-quality reward models and visual preference datasets enriched with extensive world knowledge. More critically, the RL fine-tuning process inevitably induces a distribution shift, posing an inherent risk of semantic degradation and the erosion of pre-trained knowledge. Our approach leverages text-only corpora supplemented by anti-forgetting knowledge consolidation mechanism to effectively mitigate these limitations.

\begin{figure}[tb]
\centering
\includegraphics[width=1.\linewidth]{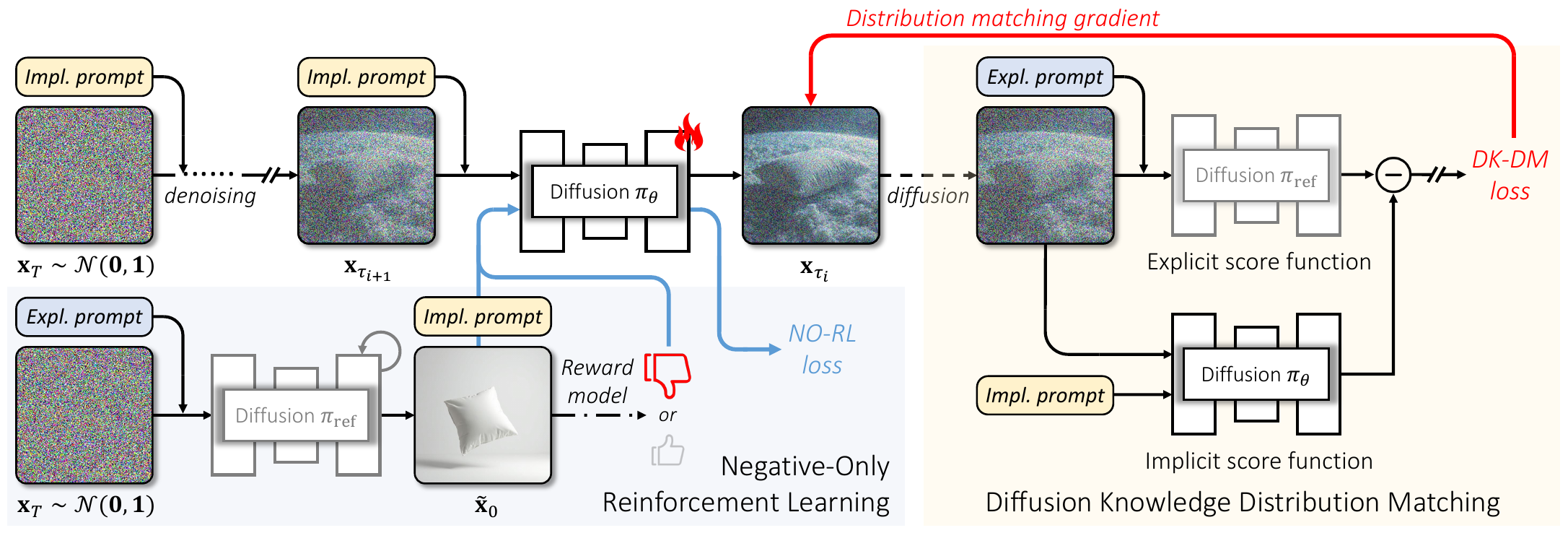}
\caption{\textbf{Illustration of \textsc{EruDiff}.} \textsc{EruDiff} focuses on the deep refactoring of the inherent knowledge systems within pre-trained text-to-image diffusion models, aiming to enhance factual fidelity when processing implicit prompts that necessitate a profound command of extensive world knowledge.}
\label{fig:3_erudiff_illustration}
\end{figure}

\section{\textsc{EruDiff}}

\textsc{EruDiff} is dedicated to the deep refactoring of the inherent knowledge systems within pre-trained diffusion models. As illustrated in Fig.~\ref{fig:3_erudiff_illustration}, \textsc{EruDiff} comprises the Diffusion Knowledge Distribution Matching (DK-DM) and the Negative-Only Reinforcement Learning (NO-RL). The former aligns the knowledge distribution of intractable implicit prompts with that of well-defined explicit anchors, while the latter rectifies inherent biases in the rendering of explicit prompts. Detailed expositions of these two components are provided in Sec.~\ref{sec:3_1} and~\ref{sec:3_2}, respectively. Sec.~\ref{sec:3_3} outlines our training strategy.

\subsection{Diffusion Knowledge Distribution Matching}
\label{sec:3_1}

Inspired by the principles of Distribution Matching Distillation (DMD) \cite{yin2024one,yin2024improved,luo2025learning}, DK-DM is implemented by aligning the synthesis of implicit prompts $\mathbf{y}_{\text{impl}}$ to that of explicit prompts $\mathbf{y}_{\text{expl}}$ at the distribution level. This is achieved by minimizing the expectation of the Kullback-Leibler (KL) divergence between the diffused implicit image distribution $p_{\text{impl},t}$ and the diffused explicit image distribution $p_{\text{expl},t}$ over the timestep $t$, both of which are derived from implicit and explicit prompts, respectively:
\begin{equation}
\begin{aligned}
\label{eq:dk-dm}
\nabla_{\mathcal{L}_{\text{DK-DM}}} &= \mathbb{E}_t\left( \nabla_\theta \mathcal{D}_{\text{KL}}\left( p_{\text{impl},t} \| p_{\text{expl},t} \right) \right)\\ &= -\mathbb{E}_{t,\mathbf{z}}\left[ \left( s_\text{expl}\left( \mathcal{F}\left( g_\theta(\mathbf{z}), t \right), t \right) - s_\text{impl}\left( \mathcal{F}\left( g_\theta(\mathbf{z}), t \right), t \right) \right) \frac{\partial g_\theta(\mathbf{z})}{\partial \theta} \right],
\end{aligned}
\end{equation}
where $t \sim \mathcal{U}(T_\text{min}, T_\text{max})$. Following \cite{yin2024one}, $T_\text{min}=0.02T$ and $T_\text{max}=0.98T$, where $T$ denotes the total number of timesteps in the training phase. $\mathbf{z} \sim \mathcal{N}(0, \mathbf{I})$ represents the random Gaussian noise input. $\mathcal{F}(\cdot, t)$ signifies the forward diffusion process (\emph{i.e.}, noise injection), with the noise level corresponding to the specific timestep $t$. In our implementation, $\mathcal{L}_{\text{DK-DM}}$ is computed by constructing a pseudo-MSE loss using the stop-gradient technique \cite{yin2024one,yin2024improved}. Furthermore, as illustrated in Fig.~\ref{fig:3_erudiff_illustration}, we have specifically designed the following components:
\begin{itemize}
    \item \textbf{Image generator} $g_\theta(\cdot)$: The image generator $g_\theta(\cdot)$ is the text-to-image diffusion model undergoing fine-tuning, denoted as $\pi_\theta(\cdot)$, which takes implicit prompts as input and is optimized via gradient backpropagation. Given that $g_\theta(\cdot)$ is a multi-step diffusion model, straightforward backpropagation through the entire denoising trajectory would lead to excessive memory consumption and potential overflow. To mitigate this issue, we employ a gradient truncation strategy, restricting the backpropagation to only the final denoising step.
    \item \textbf{Explicit score function} $s_{\text{expl}}(\cdot)$: The explicit score $s_{\text{expl}}(\cdot)$ denotes the score function of the diffused explicit image distribution, which is provided by the fixed, pre-trained text-to-image diffusion model $\pi_{\text{ref}}(\cdot, \mathbf{y}_{\text{expl}})$ conditioned on explicit prompts. Notably, the image distribution corresponding to the explicit prompts serves as the target distribution in this work.
    \item \textbf{Implicit score function} $s_{\text{impl}}(\cdot)$: The implicit score $s_{\text{impl}}(\cdot)$ denotes the score function of the diffused implicit image distribution. Distinct from \cite{yin2024one,yin2024improved,luo2025learning} that incur additional computational overhead to train a dynamically-learned denoiser, such a denoiser exists inherently within our framework in the form of the text-to-image diffusion model undergoing fine-tuning. Consequently, $s_{\text{impl}}(\cdot)$ is directly provided by the diffusion model undergoing fine-tuning $\pi_\theta(\cdot,\mathbf{y}_{\text{impl}})$ conditioned on implicit prompts.
\end{itemize}

\noindent \textbf{Full-timestep distribution matching.} In contrast to single-step \cite{yin2024one} and few-step \cite{luo2025learning} image generators, the latter of which necessitate manual alignment of denoising timesteps with the target diffusion model. In this work, the image generator and the explicit score function share an identical timestep schedule. This inherent consistency naturally motivates the execution of full-timestep distribution matching, which emerges as a more efficient alternative. Specifically, given the inference timestep sequence $(T, \dots, \tau_{i+1}, \tau_i, \tau_{i-1}, \dots, 0)$, for each training iteration, rather than exclusively selecting timestep $0$ to obtain a fully denoised clean image, we sample an intermediate timestep $\tau_i$ to generate a noisy image, upon which distribution matching is performed. 

Since our image generator $g_\theta(\cdot)$ is a multi-step diffusion model, producing a fully denoised image at every iteration is computationally prohibitive. In addition, despite the application of a gradient truncation strategy, in scenarios with a high number of inference steps, imposing constraints solely on the final denoising step exerts a relatively negligible influence on the earlier stages of the denoising chain. Consequently, performing full-timestep distribution matching serves as a natural and effective choice to address these issues. Building upon this rationale, we reformulate $\mathcal{L}_\text{DK-DM}$ as follows:
\begin{equation}
\begin{aligned}
\label{eq:dk-dm-1}
\nabla_{\mathcal{L}_{\text{DK-DM}}} = -\mathbb{E}_{t,\tau_i,z}\left[ 
\begin{aligned}
\left( s_\text{expl}\left( \mathcal{F}\left( \mathbf{x}_{\tau_i}, t \right), t \right) - s_\text{impl}\left( \mathcal{F}\left( \mathbf{x}_{\tau_i}, t \right), t \right) \right) \frac{\partial \mathbf{x}_{\tau_i}}{\partial \theta} 
\end{aligned}
\right],
\end{aligned}
\end{equation}
where $\mathbf{x}_{\tau_i} = g_\theta(\mathbf{z}, \tau_i)$, denoting that the denoising chain is truncated at the inference timestep $\tau_i$. Accordingly, $t \sim \mathcal{U}(T_\text{min}, T_\text{max})$, with $T_\text{min}=\tau_i + 0.02\times(\tau_{i+1} - \tau_i)$ and $T_\text{max}=\tau_i + 0.98\times(\tau_{i+1} - \tau_i)$. 

\noindent \textbf{Anti-forgetting knowledge consolidation.} Distribution shift is an inevitable and challenging issue encountered during the post-training of text-to-image diffusion models, \emph{e.g.}, reinforcement learning, posing a significant risk of pre-trained knowledge degradation. This risk is particularly pronounced in the context of knowledge refactoring. As illustrated in Fig.~\ref{fig:5_ablation_study_anti_forgetting}, a naive refactoring of knowledge within the diffusion model, aiming to enable it to comprehend the world-knowledge fact that “Albert Einstein's favorite musical instruments” is “Violin”, unfortuantely results in the degradation of its inherent knowledge of “Albert Einstein”. To mitigate this issue, we introduce an anti-forgetting knowledge consolidation mechanism.

Specifically, we extend the vanilla Diffusion Knowledge Distribution Matching, denoted as $\text{DK-DM}(\mathbf{y}_\text{impl}, \mathbf{y}_\text{expl})$, by incorporating an explicit knowledge consolidation term $\text{DK-DM}(\mathbf{y}_\text{expl}, \mathbf{y}_\text{expl})$ and a foundational knowledge consolidation term $\text{DK-DM}(\mathbf{y}_\text{found}, \mathbf{y}_\text{found})$, where $\mathbf{y}_\text{found}$ represents a sub-prompt extracted from the implicit prompt, specifically designed to eliminate requirements for world-knowledge reasoning. Furthermore, to enhance generalizability across diverse and complex prompts while reducing implementation difficulty, we extract {noun phrases} from the implicit prompts as $\mathbf{y}_\text{found}$, representing a more pragmatic and cost-effective alternative. For the aforementioned example, $\mathbf{y}_\text{found} \in \{\text{“Albert Einstein”, “musical instruments”}\}$. More details regarding the extraction of $\mathbf{y}_\text{found}$ are provided in the supplementary material.

Empirically, executing the anti-forgetting knowledge consolidation at a lower frequency is sufficient to effectively mitigate the loss of pre-trained knowledge. Specifically, during each training iteration, we execute $\text{DK-DM}(\mathbf{y}_\text{impl}, \mathbf{y}_\text{expl})$, $\text{DK-DM}(\mathbf{y}_\text{expl}, \mathbf{y}_\text{expl})$, and $\text{DK-DM}(\mathbf{y}_\text{found}, \mathbf{y}_\text{found})$ with probabilities $p_\text{impl}$, $p_\text{expl}$, and $p_\text{found}$, respectively. We set $p_\text{impl} = 0.8$, $p_\text{expl} = 0.1$, and $p_\text{found} = 0.1$.

\noindent \textbf{Timestep-aware curriculum learning.} Within our framework, while uniformly sampling $\tau_i$ from the inference timestep set $\{T, \dots, \tau_{i+1}, \tau_i, \tau_{i-1}, \dots, 0\}$ is a viable approach for full-timestep distribution matching, it overlooks timestep-aware characteristic, leading to suboptimal convergence efficiency. In practice, we observe that the early stages of the denoising inference chain predominantly govern the overall distribution matching. This dominance arises not only from their inherently stronger prompt-dependency \cite{balaji2022ediff} but also from the cascading impact on the matching of subsequent timesteps. Consequently, we employ a timestep-aware curriculum learning strategy to prioritize these timesteps:
\begin{equation}
\begin{aligned}
\label{eq:ta-cl}
p(n) = \frac{\exp(-\lambda (n - 1))}{\sum_{i=1}^{T_\text{inference}} \exp(-\lambda (i - 1))}, 
\end{aligned}
\end{equation}
where $n$ is the number of inference steps performed, $n \in \{1, 2, \dots, T_\text{inference}\}$, $T_{\text{inference}}$ represents the total number of timesteps in the inference phase, and $p(n)$ represents the probability that the image generator $g_\theta(\cdot)$ performs only the first $n$ denoising operations within the current training iteration. $\lambda$ represents the decay coefficient, which is set to $\lambda = 0.1$ in this work. Values in the range of $0.05$ to $0.2$ represent moderate and effective choices.

\subsection{Negative-Only Reinforcement Learning}
\label{sec:3_2}

While state-of-the-art text-to-image diffusion models demonstrate a strong command of explicit prompts, they inevitably exhibit inherent representation biases and inaccurate visual rendering to some extent (see Fig.~\ref{fig:5_ablation_study_no_rl}). To address these limitations, we employ Negative-Only Reinforcement Learning (NO-RL) for fine-grained correction, thereby achieving superior generative outcomes.

In this work, NO-RL is implemented as a variant of Kahneman-Tversky Optimization (KTO) \cite{ethayarajh2024kto}, which offers the distinct advantage of bypassing the need for costly paired preference data, instead, it aligns the model with human preferences using only binary feedback signals, making it inherently compatible with our problem formulation. As illustrated in Fig.~\ref{fig:3_erudiff_illustration}, we exclusively leverage the failure sample set $\mathcal{D}_\text{loss}$, derived from explicit prompt synthesis, for the NO-RL process. Specifically, following the methodology in \cite{li2024aligning}, the image generator $g_\theta(\cdot)$ is further optimized using the following objective function:
\begin{equation}
\begin{aligned}
\label{eq:no-rl}
\max_{\pi_{\theta}}\mathbb{E}_{\tilde{\mathbf{x}}_0, t}\left[ U\left(-\left( \beta\log\frac{\pi_\theta(\mathbf{x}_{t-1}|\mathbf{x}_t)}{\pi_\text{ref}(\mathbf{x}_{t-1}|\mathbf{x}_t)} - Q_\text{ref} \right)\right) \right],
\end{aligned}
\end{equation}
where $\tilde{\mathbf{x}}_0 \in \mathcal{D}_{\text{loss}}$, $t \sim \mathcal{U}[0, T]$, $\mathbf{x}_t=\mathcal{F}(\tilde{\mathbf{x}}_0,t)$. $\pi_*(x_{t-1} | x_t)$ represents a denoising sampling step. $U(\cdot)$ denotes a monotonically increasing value function that maps implicit rewards to subjective utilities, which is instantiated as a sigmoid function in this work. The reference point $Q_{\text{ref}}$ is computed by evaluating the term $\max(0, \frac{1}{m} \sum \beta \log \frac{\pi_{\theta}(\mathbf{x}_{t-1}'|\mathbf{x}_{t}')}{\pi_{\text{ref}}(\mathbf{x}_{t-1}'|\mathbf{x}_{t}')})$ over a batch of $m$ unrelated $(\mathbf{x}_{t-1}', \mathbf{x}_{t}')$ pairs.

NO-RL effectively reshapes the target distribution by excluding the distribution space associated with failure samples. Given that the positive distribution representation is already sufficiently captured by DK-DM, NO-RL adopts a more efficient strategy by focusing exclusively on constraints derived from negative samples, the integration of redundant positive sample learning into the existing NO-RL framework yields no further performance gains. Further methodological details, including the filtering strategy for the failure sample set $\mathcal{D}_\text{loss}$, are provided in the supplementary material.

\begin{algorithm}[tb]
\caption{\textsc{EruDiff}}
\label{alg:erudiff}
\begin{algorithmic}[1]
\REQUIRE Pre-trained text-to-image diffusion model weights $\theta$, prompt dataset $\mathcal{Y}$, sampling priors $\{p_{\text{impl}}, p_{\text{expl}}, p_{\text{found}}\}$, failure sample set $\mathcal{D}_{\text{loss}}$.\\
\WHILE{not converged}
    \STATE \textit{// Phase I: Diffusion Knowledge Distribution Matching (DK-DM)}
    \STATE Sample prompt category $c \in \{\text{impl, expl, found}\}$ with probability $p_c$;
    \STATE Sample prompt triplet $(\mathbf{y}_\text{impl}, \mathbf{y}_\text{expl}, \mathbf{y}_\text{found}) \in \mathcal{Y}$;
    \STATE Construct corresponding prompt pair;
    \STATE \quad \textbf{if} $c$ \textbf{is} impl \textbf{then} $(\mathbf{y}_\text{impl}, \mathbf{y}_\text{expl}) \leftarrow (\mathbf{y}_\text{impl}, \mathbf{y}_\text{expl})$;
    \STATE \quad \textbf{if} $c$ \textbf{is} expl \textbf{then} $(\mathbf{y}_\text{impl}, \mathbf{y}_\text{expl}) \leftarrow (\mathbf{y}_\text{expl}, \mathbf{y}_\text{expl})$;
    \STATE \quad \textbf{if} $c$ \textbf{is} found \textbf{then} $(\mathbf{y}_\text{impl}, \mathbf{y}_\text{expl}) \leftarrow (\mathbf{y}_\text{found}, \mathbf{y}_\text{found})$;
    \STATE Sample Gaussian noise $\mathbf{z} \sim \mathcal{N}(0, \mathbf{I})$ and step count $n \sim p(n)$ using Eq.~\ref{eq:ta-cl};
    \STATE Execute $n$ denoising steps to obtain truncated latent $\mathbf{x}_{\tau_i} = g_\theta(\mathbf{z}, \tau_i)$;
    \STATE Sample $t \sim \mathcal{U}(T_{\text{min}}, T_{\text{max}})$ relative to interval $[\tau_i, \tau_{i+1}]$;
    \STATE Calculate ${\mathcal{L}_{\text{DK-DM}}}$ using Eq. \ref{eq:dk-dm-1};
    \STATE Perform EMAN: ${\mathcal{L}_{\text{DK-DM}}} \leftarrow \text{EMAN}(\mathcal{L}_{\text{DK-DM}})$;
    \STATE Update $\theta \leftarrow \theta - \eta \cdot \nabla_{\mathcal{L}_{\text{DK-DM}}}$;

    \STATE \textit{// Phase II: Negative-Only Reinforcement Learning (NO-RL)}
    \STATE Sample failure sample $\tilde{\mathbf{x}}_0 \sim \mathcal{D}_{\text{loss}}$ and timestep $t \sim \mathcal{U}[0, T]$;
    \STATE Compute ${\mathcal{L}_{\text{NO-RL}}}$ using Eq. \ref{eq:no-rl};
    \STATE Perform EMAN: ${\mathcal{L}_{\text{NO-RL}}} \leftarrow \text{EMAN}(\mathcal{L}_{\text{NO-RL}})$;
    \STATE Update $\theta \leftarrow \theta - \eta \cdot \nabla_{\mathcal{L}_{\text{NO-RL}}}$;
\ENDWHILE
\end{algorithmic}
\end{algorithm}

\subsection{Training strategy}
\label{sec:3_3}

As illustrated in Fig.~\ref{fig:3_erudiff_illustration}, the optimization objective of our method comprises two components: DK-DM and NO-RL. Although both components aim to optimize the image generator $g_{\theta}(\cdot)$, they do not share the same network forward pass within the computational graph, distinguishing them from standard multi-objective joint training paradigms. Consequently, we perform two independent backpropagation steps to facilitate the learning

This strategy effectively reduces peak memory consumption without incurring additional computational overhead from redundant forward passes. Furthermore, to balance the two optimization objectives that operate at different scales, we implement an exponential moving average normalization (EMAN) mechanism with a decay coefficient of 0.99 for both loss functions to ensure training robustness and convergence stability. Algorithm~\ref{alg:erudiff} outlines the final training procedure. 

\begin{figure}[tb]
\centering
\includegraphics[width=1.\linewidth]{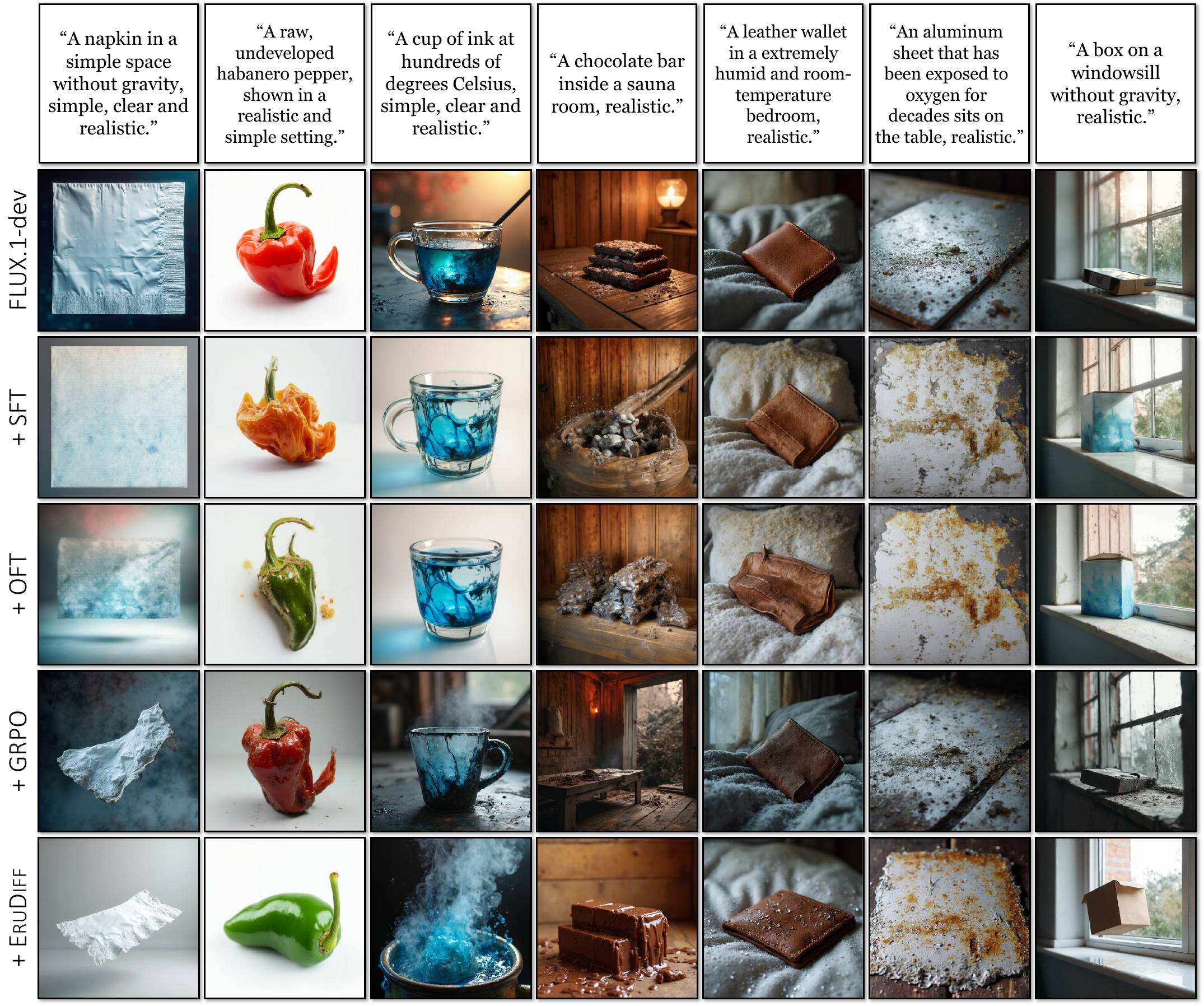}
\caption{\textbf{Qualitative comparison on \textsc{Science}-T2I based on FLUX.} Each image is generated with the same text prompt and random seed for all methods. \textsc{EruDiff} excels in precise scientific knowledge comprehension and rendering.}
\label{fig:5_qualitative_comparison_flux}
\end{figure}

\section{Knowledge-10K}

Despite the prevalence of world knowledge evaluation benchmarks, there remains a critical lack of structured training resources. To address this deficiency, we introduce the Knowledge-10K dataset for training support, which will be made available to the public to foster further investigation within the community.

\noindent \textbf{Taxonomy.} Adhering to the design principle of prevailing evaluation benchmark \cite{niu2025wise}, Knowledge-10K encompasses world knowledge across three primary domains: {cultural commonsense}, {spatio-temporal reasoning}, and {natural science}. Cultural commonsense spans diverse fields such as festivals, sports, and craftsmanship, it entails the recognition of traditional customs, ethnic handicrafts, and iconic landmarks, as well as events associated with prominent figures, among other cultural manifestations. Spatio-temporal reasoning integrates both temporal and spatial dimensions, necessitating reasoning regarding chronological relationships, positioning, and perspectives. Natural science incorporates domain-specific expertise across biology, physics, and chemistry.

\noindent \textbf{Format.} Each entry in the Knowledge-10K dataset primarily consists of an \emph{implicit prompt} and its corresponding \emph{explicit prompt}. The former necessitates the retrieval and reasoning of deep-seated world knowledge, while the latter serves as its intuitive counterpart, which circumvents logical complexity by directly describing the visual content. Samples are provided in the supplementary material.

\begin{figure}[tb]
\centering
\setlength{\belowcaptionskip}{-0.3cm}
\includegraphics[width=1.\linewidth]{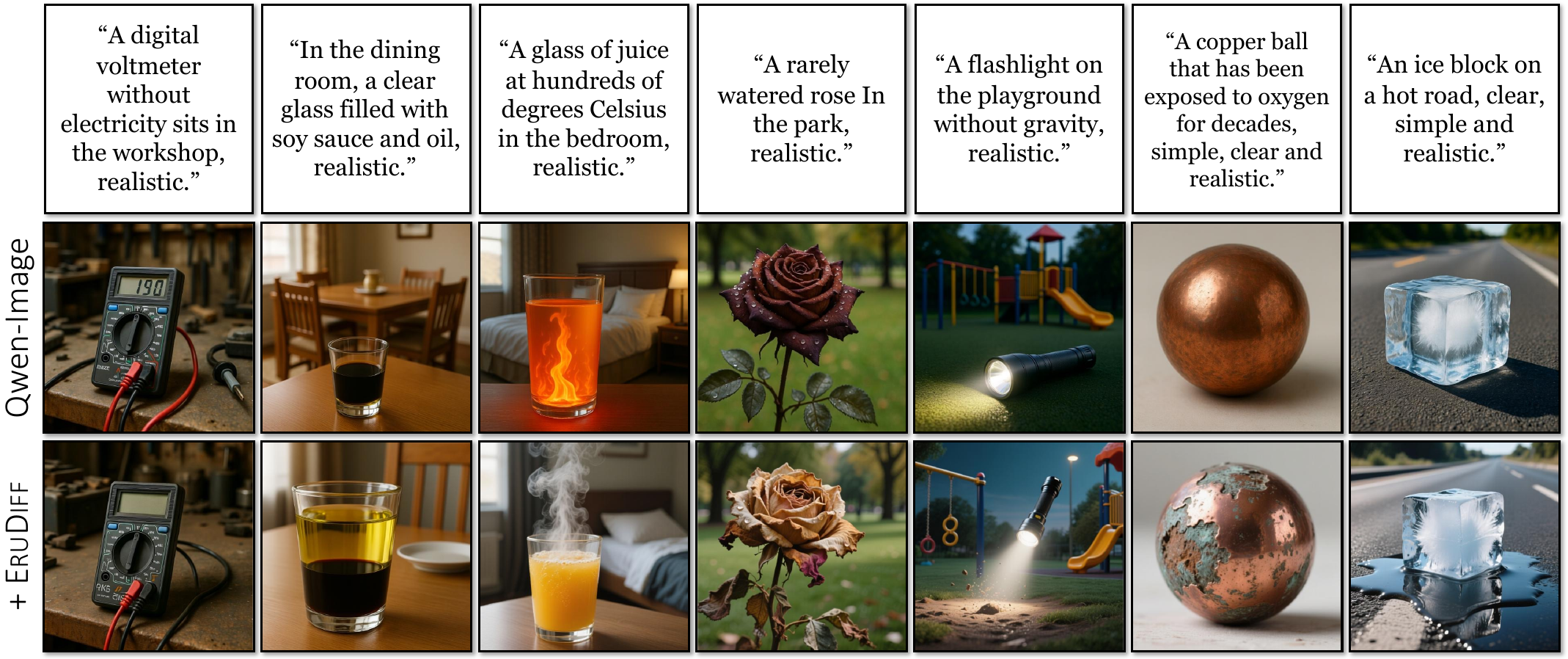}
\caption{\textbf{Qualitative comparison on \textsc{Science}-T2I based on Qwen-Image.} Each image is generated with the same text prompt and random seed for all methods. \textsc{EruDiff} excels in precise scientific knowledge comprehension and rendering.}
\label{fig:5_qualitative_comparison_qwen_image}
\end{figure}

\noindent \textbf{Data collection pipeline.} (1) Template customization. Initially, 100 seed samples were manually constructed to establish the specifications for target data characteristics and to define the taxonomy of world knowledge involved. (2) Scale expansion. We adopted a hybrid strategy combining information retrieval with Large Language Models (LLMs) synthesis. The former involves extracting raw information from the Internet and encyclopedic sources, followed by manual rewriting to ensure format alignment. The latter leverages the collaborative synergy of Gemini 3, GPT-5, and Grok 4 to synthesize new entries, which mitigates the redundancy and stylistic uniformity inherent in single-model generation. Subsequently, Gemini 3 was employed to perform semantic deduplication, filtering out redundant instances to yield an initial corpus of 10,000 entries. (3) Data review. To mitigate retrieval bias and LLM hallucinations, the more advanced Gemini 3 Pro was introduced for automated proofreading and correction. Finally, five volunteers holding at least a Bachelor's degree in Engineering were invited to conduct a final audit, thereby ensuring the high fidelity and rigorous quality of the dataset.

\noindent \textbf{Dataset Statistics.} The Knowledge-10K dataset comprises 10,000 data entries. The domains of cultural commonsense, spatio-temporal reasoning, and natural science consist of 4,000, 3,000, and 3,000 samples, respectively.

\begin{table}[tb]
\caption{\textbf{Objective evaluation on WISE.} \textsc{EruDiff} yields consistent performance gains across all categories, marking a significant advancement in the field of image synthesis guided by world knowledge. \textbf{Bold} numbers indicate the highest scores among text-to-image diffusion models without the aid of multimodal large language models.}
\label{tab:wise}
\centering
\resizebox{0.99\linewidth}{!}{
\begin{tabular}{@{}lcccccccc@{}}
% \begin{tabular}{lcccccccc}
\toprule
\multicolumn{1}{c}{\bf Methods} & \textbf{Type} & {Cultural} & {Time} & {Space} & {Biology} & {Physics} & {Chemistry} & \textbf{Overall} \\
\midrule
Janus-Pro-7B \cite{chen2025janus} & \emph{unified} & 0.30 & 0.37 & 0.49 & 0.36 & 0.42 & 0.26 & 0.35 \\
Emu3 \cite{wang2024emu3} & \emph{unified} & 0.34 & 0.45 & 0.48 & 0.41 & 0.45 & 0.27 & 0.39 \\
BLIP3o-8B \cite{chen2025blip3} & \emph{unified} & 0.49 & 0.51 & 0.63 & 0.54 & 0.63 & 0.37 & 0.52\\
BAGEL \cite{deng2025emerging} & \emph{unified} & 0.44 & 0.55 & 0.68 & 0.44 & 0.60 & 0.39 & 0.52 \\
BAGEL + CoT \cite{deng2025emerging} & \emph{unified} & 0.76 & 0.69 & 0.75 & 0.65 & 0.75 & 0.58 & 0.70 \\ 
\midrule
% \midrule
Qwen-Image \cite{wu2025qwen} & \makecell{\scriptsize \emph{MLLM} +\\\scriptsize \emph{diffusion}}  & 0.62 & 0.63 & {0.77} & {0.57} & {0.75} & 0.40 & 0.62 \\
\midrule
SD 1.5 \cite{rombach2022high} & \emph{diffusion} & 0.34 & 0.35 & 0.32 & 0.28 & 0.29 & 0.21 & 0.32 \\
SD 2.1 \cite{rombach2022high} & \emph{diffusion} & 0.30 & 0.38 & 0.35 & 0.33 & 0.34 & 0.21 & 0.32 \\
SD XL \cite{podell2024sdxl} & \emph{diffusion} & 0.43 & 0.48 & 0.47 & 0.44 & 0.45 & 0.27 & 0.43 \\
PixArt-$\alpha$ \cite{chen2024pixart} & \emph{diffusion} & 0.45 & 0.50 & 0.48 & 0.49 & 0.56 & 0.34 & 0.47 \\
SD 3-medium \cite{esser2024scaling} & \emph{diffusion} & 0.42 & 0.44 & 0.48 & 0.39 & 0.47 & 0.29 & 0.42 \\
SD 3.5-medium \cite{esser2024scaling} & \emph{diffusion} & 0.43 & 0.50 & 0.52 & 0.41 & 0.53 & 0.33 & 0.45 \\
SD 3.5-large \cite{esser2024scaling} & \emph{diffusion} & 0.44 & 0.50 & 0.58 & 0.44 & 0.52 & 0.31 & 0.46 \\
FLUX.1-schnell \cite{flux2024} & \emph{diffusion} & 0.39 & 0.44 & 0.50 & 0.31 & 0.44 & 0.26 & 0.40 \\
FLUX.1-dev \cite{flux2024} & \emph{diffusion} & 0.48 & 0.58 & 0.62 & 0.42 & 0.51 & 0.35 & 0.50 \\
% \midrule
FLUX.1-dev + \textsc{EruDiff} & \emph{diffusion} & {\bf 0.66} & {\bf 0.65} & {\bf 0.72} & {\bf 0.54} & {\bf 0.64} & {\bf 0.49} & {\bf 0.64} \\
\bottomrule
\end{tabular}}
\end{table}

\section{Experiments}

\subsection{Experimental settings}

\noindent \textbf{Implementation details.} \textsc{EruDiff} is designed as a model-agnostic training paradigm, engineered for seamless integration into diverse state-of-the-art text-to-image diffusion models. To rigorously evaluate its performance and cross-model generalizability, we instantiate \textsc{EruDiff} upon two representative high-capacity backbones: FLUX.1-dev \cite{flux2024} and Qwen-Image \cite{wu2025qwen}. Our implementation is built upon the Diffusers library, utilizing Low-Rank Adaptation (LoRA) to efficiently fine-tune the attention layers within the denoising network. More parameter setting, training details are provided in the supplementary material.

\noindent \textbf{Benchmarks.} The performance of \textsc{EruDiff} is rigorously assessed across two widely recognized benchmarks: scientific knowledge benchmark (\emph{i.e.}, \textsc{Science}-T2I \cite{li2025science}) and the world knowledge benchmark (\emph{i.e.}, WISE \cite{niu2025wise}). 

\textsc{Science}-T2I \cite{li2025science} spans multiple scientific domains, including physics, chemistry, and biology, encompassing 16 distinct scientific phenomena. It introduces \textsc{SciScore}, an end-to-end reward model integrated with expert-level scientific knowledge, designed to evaluate whether generated images precisely reflect the scientific phenomena described in the given prompts. The \textsc{Science}-T2I dataset comprises both training and testing sets, the latter is further divided into \textsc{Science}-T2I S and \textsc{Science}-T2I C. Specifically, \textsc{Science}-T2I S maintains the same stylistic attributes as the training set, whereas \textsc{Science}-T2I C introduces more diverse scene settings. We evaluate the performance of the proposed method using the corresponding dataset splits.

WISE \cite{niu2025wise} is a benchmark specifically designed for world knowledge-informed semantic evaluation. It utilizes 1,000 meticulously crafted prompts to assess the capability of text-to-image models in understanding and rendering world knowledge across 25 subdomains in cultural commonsense, spatio-temporal reasoning, and natural science. Correspondingly, WISE introduces WiScore, a quantitative metric for evaluating world knowledge-image alignment. As WISE serves solely as an evaluation benchmark and does not provide associated training resources, We utilize the constructed Knowledge-10K dataset as the training support.

\begin{table}[tb]
\caption{\textbf{Objective evaluation on \textsc{Science}-T2I.} \textsc{EruDiff} outperforms SFT and RL-based strategies and MLLM-dependent framework, demonstrating substantial improvements in scientific knowledge informed image synthesis. The best performance is highlighted with \textbf{bold} values.}
\label{tab:science-t2i}
\centering
\resizebox{0.78\linewidth}{!}{
\begin{tabular}{@{}lc@{\quad}c@{}}
% \begin{tabular}{lcccccccc}
\toprule
\multicolumn{1}{c}{\bf Methods} & \makecell{\bf Science-T2I S\\ \textsc{SciScore}} & \makecell{\bf Science-T2I C\\ \textsc{SciScore}} \\
\midrule
FLUX.1-dev \cite{flux2024} & 23.31 & 27.54 \\
FLUX.1-dev + prompt rewrite (GPT-4o) & 32.90 & 33.19 \\
FLUX.1-dev + SFT \cite{li2025science} & 28.64 & 30.01 \\
FLUX.1-dev + OFT \cite{li2025science} & 30.95 & 32.31 \\
FLUX.1-dev + GRPO \cite{liu2025flow} & 31.68 & 32.19 \\
% \midrule
FLUX.1-dev + \textsc{EruDiff} (Ours) & \textbf{33.70} & \textbf{34.54} \\
% \bottomrule
% \toprule
% \multicolumn{1}{c}{\bf Methods} & \textbf{\textsc{Science-T2I S}} & \textbf{\textsc{Science-T2I C}} \\
\midrule
Qwen-Image \cite{wu2025qwen} & 25.38 & 31.56 \\
Qwen-Image + prompt rewrite (GPT-4o) & 34.52 & 37.86 \\
Qwen-Image + \textsc{EruDiff} (Ours) & \textbf{35.43}  & \textbf{38.48} \\
\bottomrule
\end{tabular}
}
\end{table}

\subsection{Qualitative comparison}

Fig.~\ref{fig:5_qualitative_comparison_flux}, \ref{fig:5_qualitative_comparison_qwen_image} present a comparative analysis of our results against state-of-the-art counterparts on the \textsc{Science}-T2I benchmark based on FLUX \cite{flux2024} and Qwen-Image \cite{wu2025qwen}, respectively. SFT involves supervised fine-tuning using high-quality text-image pairs precisely rendered with scientific knowledge from \textsc{Science}-T2I. OFT \cite{li2025science}, an enhanced variant of DPO \cite{rafailov2023direct}, and GRPO \cite{liu2025flow} are both reinforcement learning (RL) based strategies trained via the \textsc{SciScore} reward model provided by \textsc{Science}-T2I.

As shown in Fig.~\ref{fig:5_qualitative_comparison_flux}, although these methods marginally improve the performance of the pre-trained text-to-image diffusion model (\emph{i.e.}, FLUX.1-dev), they continue to struggle with unnatural renderings (\emph{e.g.}, “stiff floating of napkin without gravity”), as well as the omission of critical elements (\emph{e.g.}, “chocolate”) and unintended attribute leakage (\emph{e.g.}, “erroneous attachment of blue tints to napkin and box”). In contrast, \textsc{EruDiff} yields more scientifically accurate and visually realistic synthesis, outperforming existing counterparts. Notably, \textsc{EruDiff} is trained exclusively on the text resources from \textsc{Science}-T2I, offering superior scalability compared to SFT and RL-based approaches, for which the requisite high-quality text-image pairs and reward models are often prohibitively expensive. Furthermore, as shown in Fig.~\ref{fig:5_qualitative_comparison_qwen_image}, even though Qwen-Image incorporates a multimodal large language model (\emph{i.e.}, Qwen2.5-VL) as a text encoder to derive superior representations, it still fails to capture complex phenomena, such as “immiscible liquid mixing”, “wilting states”, and “corrosion”. \textsc{EruDiff} successfully masters these concepts, thereby enhancing overall performance.

Fig.~\ref{fig:5_qualitative_comparison_flux_wise} presents visual results on the WISE benchmark, \textsc{EruDiff} demonstrates a robust command of world knowledge, ranging from natural science to cultural commonsense, resulting in significant performance enhancements.

\begin{figure}[tb]
\centering
\includegraphics[width=1.\linewidth]{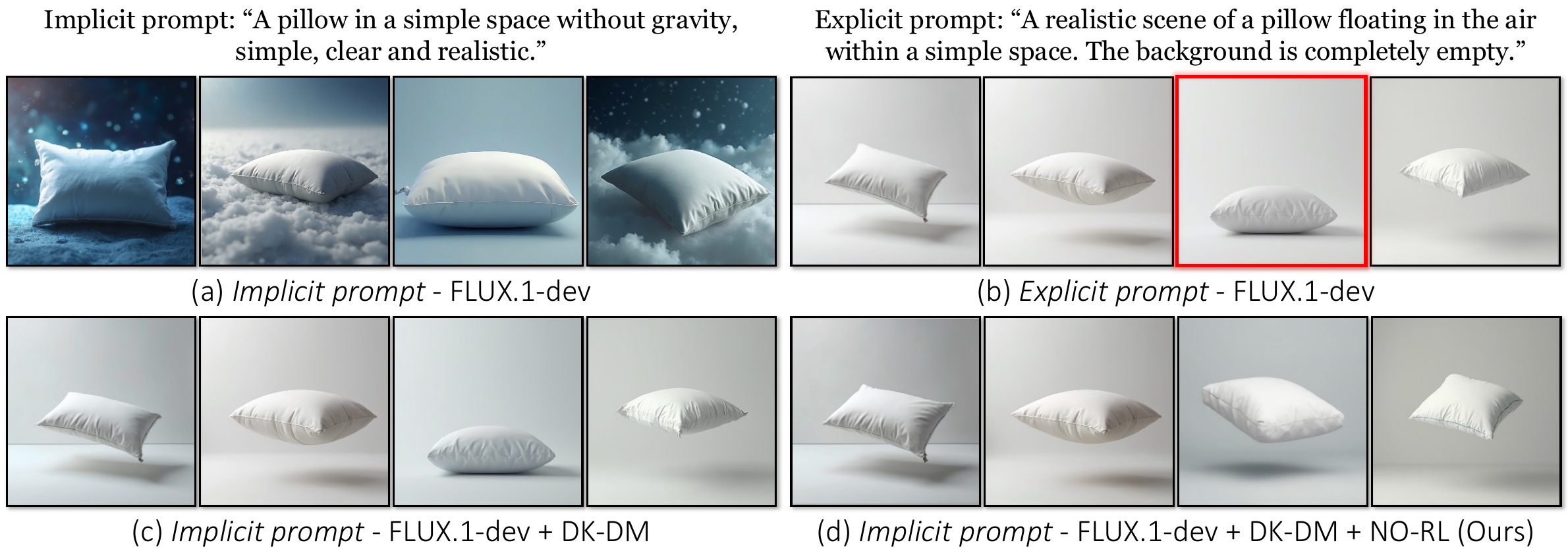}
\caption{\textbf{Ablation study} on negative-only reinforcement learning, which effectively rectifies inherent biases in explicit prompt rendering, as highlighted by the red box.}
\label{fig:5_ablation_study_no_rl}
\end{figure}

\subsection{Quantitative comparison}

\noindent \textbf{Objective evaluation.} Tables~\ref{tab:wise} and~\ref{tab:science-t2i} present the objective evaluations conducted on the WISE and \textsc{Science}-T2I benchmarks, respectively. As indicated in Table~\ref{tab:wise}, \textsc{EruDiff} fully unleashes the world knowledge informed image synthesis capabilities of FLUX, significantly enhancing performance across all evaluation dimensions. It surpasses Qwen-Image, which leverages multimodal large language model (\emph{i.e.}, Qwen2.5-VL) capabilities, and achieves state-of-the-art performance among diffusion-based models. Furthermore, the objective evaluation on the scientific knowledge benchmark in Table~\ref{tab:science-t2i} demonstrates that our method outperforms both SFT and RL-based strategies. Notably, by effectively rectifying the inherent biases in explicit prompt rendering, our approach also demonstrates superior performance compared to prompt rewriting using the advanced commercial model GPT-4o, despite the capacity of latter for scientific knowledge comprehension.

\noindent \textbf{User study.} A subjective user study is provided in the supplementary material.

\subsection{Ablation study}

\noindent \textbf{On negative-only reinforcement learning.} As illustrated in Fig.~\ref{fig:5_ablation_study_no_rl}, NO-RL focuses on circumventing the rendering inaccuracies inherently present in existing text-to-image diffusion models when processing explicit prompts, highlighted by the red box. These inaccuracies are otherwise propagated to the fine-tuned model following DK-DM. By leveraging these failure samples for single-preference reinforcement learning, NO-RL effectively mitigates this issue.

\noindent \textbf{On anti-forgetting knowledge consolidation.} As shown in Fig.~\ref{fig:5_ablation_study_anti_forgetting}, a naive implementation of knowledge refactoring inevitably leads to the degradation of pre-trained knowledge, such as the erosion of the “Albert Einstein” concept. By introducing AF-KC, we achieve effective preservation of these knowledge.

\noindent \textbf{On timestep-aware curriculum learning.} Given the timestep-aware characteristic of DK-DM, we design a customized timestep-aware curriculum learning strategy that prioritizes the early stages of the denoising inference chain. As shown in Fig.~\ref{fig:5_ablation_study_timestep_cl}, our method achieves accelerated convergence compared to the standard uniform timestep sampling baseline.

\begin{table}[tb]
    \centering
    \begin{minipage}[t]{0.57\textwidth}
        \centering
        \includegraphics[width=1.\textwidth]{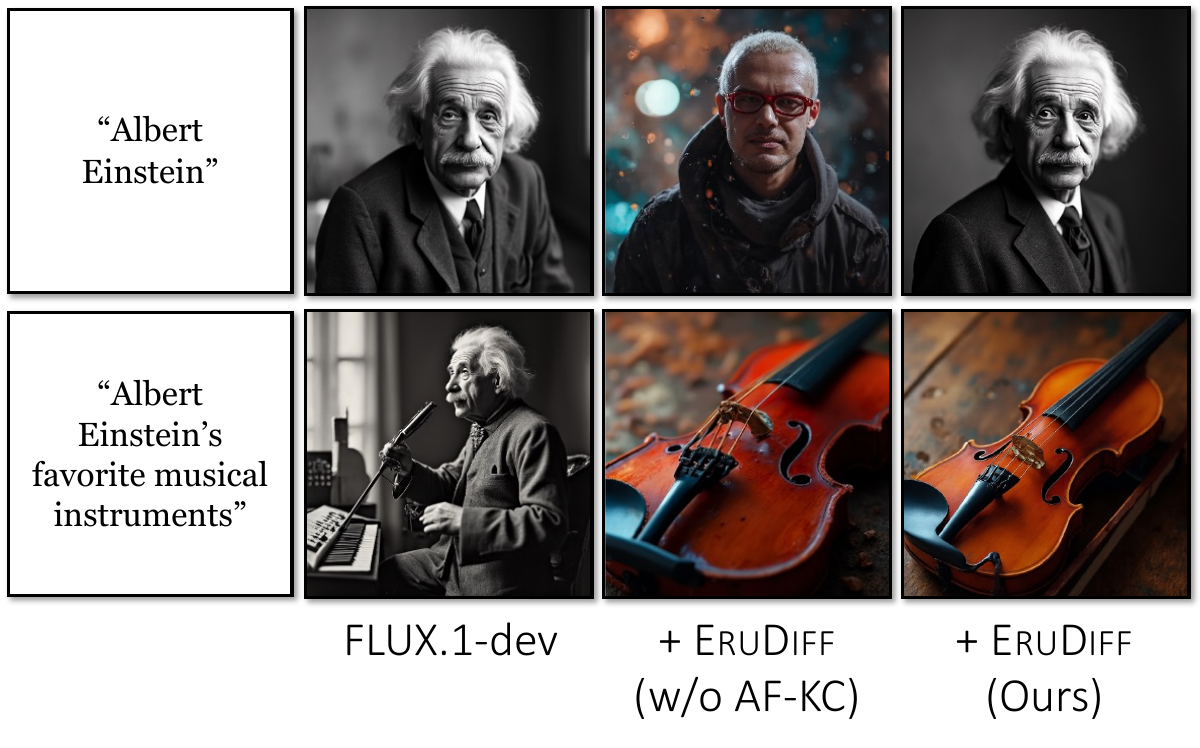}
        \captionof{figure}{\textbf{Ablation study} on AF-KC.}
        \label{fig:5_ablation_study_anti_forgetting}
    \end{minipage}
    \hfill
    \begin{minipage}[t]{0.42\textwidth}
        \centering
        \includegraphics[width=1.\textwidth]{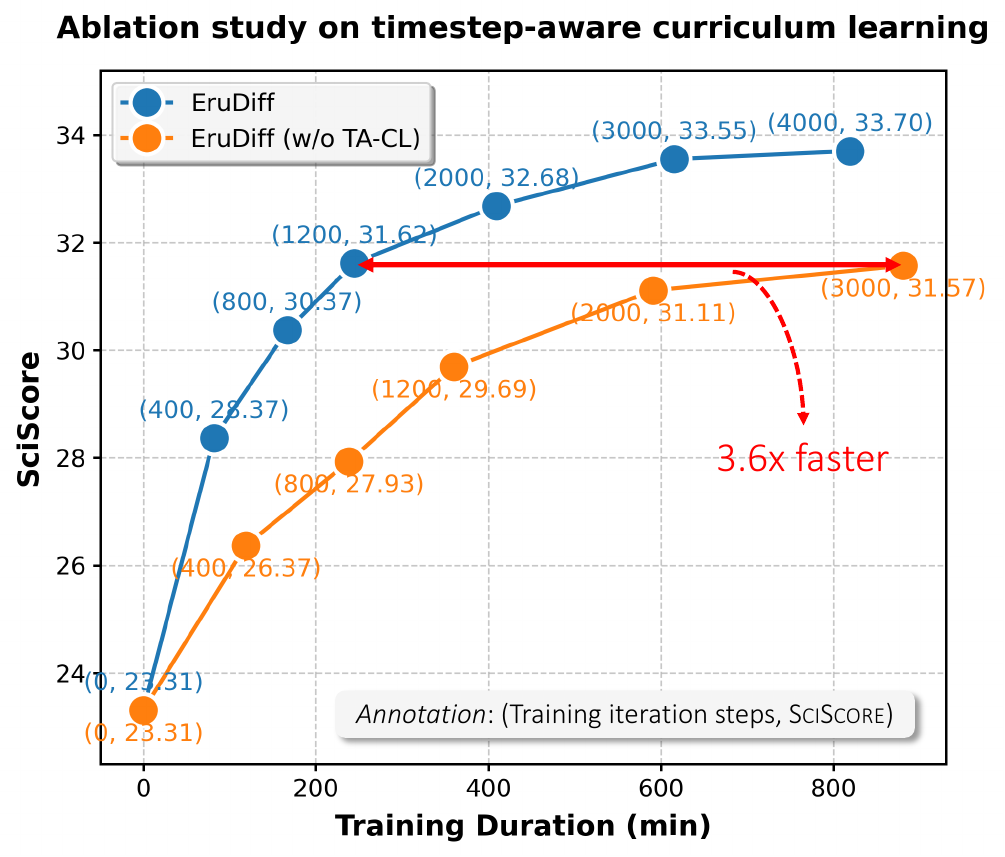}
        \captionof{figure}{\textbf{Ablation study} on TA-CL.}
        \label{fig:5_ablation_study_timestep_cl}
    \end{minipage}
\end{table}

\section{Conclusion}

This study focuses on the crucial shortcoming in text-to-image diffusion models concerning their inability to generate factually accurate images from implicit prompts requiring profound world knowledge, despite excelling in fidelity with explicit prompts. The proposed EruDiff mitigates this issue through Diffusion Knowledge Distribution Matching (DK-DM), which registers the distributions of implicit prompts with well-defined explicit anchors, and is further complemented by Negative-Only Reinforcement Learning (NO-RL) to precisely rectify inherent biases in explicit prompt rendering, ultimately leading to enhanced performance. Rigorous empirical evaluations on the \textsc{Science}-T2I and WISE benchmarks validate that EruDiff significantly enhances the performance of leading models, including FLUX and Qwen-Image, underscoring its efficacy in fostering more factually consistent and scientifically grounded generative AI.

% \section*{Acknowledgements}
% Please insert your acknowledgments here.

\bibliographystyle{splncs04}
\bibliography{main}

\clearpage

\appendix

This appendix is structured as follows:

\begin{itemize}
    \item In Appendix~\ref{sec:A}, we provide further implementation details, including training configurations, data preprocessing pipelines, and essential methodological specifications.
    \item In Appendix~\ref{sec:B}, we provide comprehensive details regarding the construction of the Knowledge-10K dataset, including data formats and instruction templates.
    \item In Appendix~\ref{sec:C}, we provide additional experimental results, including user study, additional ablation analyses, and additional visual results.
\end{itemize}

\section{More implementation details}
\label{sec:A}

\noindent \textbf{Training details.} The performance of \textsc{EruDiff} is rigorously evaluated across two widely recognized benchmarks: the scientific knowledge benchmark (\textit{i.e.}, \textsc{Science-T2I} \cite{li2025science}) and the world knowledge benchmark (\textit{i.e.}, WISE \cite{niu2025wise}).

For the \textsc{Science-T2I} benchmark, we instantiate \textsc{EruDiff} using Low-Rank Adaptation (LoRA) on two representative state-of-the-art backbones: FLUX.1-dev \cite{flux2024} and Qwen-Image \cite{wu2025qwen}. For both models, we employ the AdamW optimizer with a learning rate of $5\times10^{-5}$. The optimizer hyperparameters are set to $\beta_1=0.9$, $\beta_2=0.999$, $\epsilon=1\times10^{-8}$, with a weight decay of $1\times10^{-4}$. The LoRA is applied to the attention layers of the denoising network, and the rank is set to 16. For FLUX.1-dev \cite{flux2024}, the fine-tuning process is conducted on 8 NVIDIA A100 GPUs with a batch size of 2 per device and a gradient accumulation step of 4, resulting in an effective batch size of 64. The fine-tuning is completed with 4,000 training steps. Consistent with \cite{li2025science}, the number of inference timesteps and the guidance scale are set to 30 and 0, respectively. For Qwen-Image \cite{wu2025qwen}, the fine-tuning process is performed on 8 NVIDIA A100 GPUs with a batch size of 1 per device and a gradient accumulation step of 8, yielding an effective batch size of 64. Fine-tuning is conducted for 5,000 optimization steps. Following default configurations, the number of inference timesteps is set to 50, and the guidance scale is set to 4.

For the WISE benchmark, we instantiate \textsc{EruDiff} on FLUX.1-dev \cite{flux2024} using LoRA. Training is performed on the Knowledge-10K dataset. We utilize the AdamW optimizer with a learning rate of $5\times10^{-5}$. The optimizer hyperparameters are identical to those used in the \textsc{Science-T2I} experiments, \textit{i.e.}, $\beta_1=0.9$, $\beta_2=0.999$, $\epsilon=1\times10^{-8}$, and weight decay of $1\times10^{-4}$. The LoRA is applied to the attention layers of the denoising network, and the rank is set to 128. Training is conducted on 8 NVIDIA A100 GPUs with a batch size of 2 per device and a gradient accumulation step of 4, totaling an effective batch size of 64. Fine-tuning is conducted for 50,000 optimization steps. The number of inference timesteps is set to 28, and the guidance scale is set to 3.5.

\noindent \textbf{Workflow for $\mathbf{y}_\text{found}$ Extraction.} Regarding the workflow for the extraction of $\mathbf{y}_{\text{found}}$, as mentioned in Sec. 3.1, we extract noun phrases from implicit prompts to serve as $\mathbf{y}_{\text{found}}$. This approach offers a more pragmatic and cost-effective alternative, enhancing generalization capabilities and reducing implementation complexity when handling diverse and intricate prompts. In our experiments, we utilize DeepSeek-V3.2 for the preprocessing task of extracting $\mathbf{y}_{\text{found}}$, the corresponding instructions are provided as follows:

\begin{tcolorbox}[
    colback=MorandiBlue, 
    colframe=MorandiBlueFrame, 
    arc=4pt, 
    boxrule=1pt, 
    left=5pt, right=5pt, top=5pt, bottom=5pt, 
    % title=A,
]
\small{
You extract only \textbf{entity nouns} and \textbf{proper nouns} from the given prompt.\\
\textbf{Task:}\\
Given the input \textit{prompt}, return \textit{sub\_prompt} as a concise list of:\\
1. \textbf{Entity nouns} (concrete objects, animals, plants, tools, places, etc.).\\
2. \textbf{Proper nouns} (person names, place names, organizations, etc.).
\\
\textbf{Extraction policy:}\\
1. Keep wording from the original prompt whenever possible.\\
2. Prefer short noun units (\textit{e.g.}, “A bird”, “The shield”, “Captain America”).\\
3. If there are multiple valid entities, include multiple items.\\
4. If uncertain, prefer dropping ambiguous items.\\
\textbf{Must NOT include:}\\
1. Abstract nouns (\textit{e.g.}, “invention”, “moment”, “symbol”, “concept”, etc.).\\
2. Verbs, clauses, or full descriptive sentences.\\
3. Pure adjectives, adverbs.\\
\textbf{Examples:}\\
Input: “The shield used by Captain America”\\
Output: [“The shield”, “Captain America”]\\
Input: “The sword of King Arthur that he pulled from the stone”\\
Output: [“The sword”, “King Arthur”, “stone”]\\
\\
Here is the input prompt: [input prompt]
}
\end{tcolorbox}

\noindent \textbf{Filtering strategy for $\mathcal{D}_{loss}$.} For the \textsc{Science-T2I} benchmark, To maintain consistency with SFT and RL-based strategies, we employ \textsc{SciScore} for the filtering of failure samples. As \textsc{SciScore} provides a continuous score rather than a direct binary evaluation for a given input image, we adopt a simple yet effective thresholding strategy. Specifically, we synthesize 1,000 images using a pre-trained text-to-image diffusion model based on randomly sampled implicit prompts, and we define the mean of their corresponding scores as the threshold. Samples with scores below this threshold are categorized into the failure sample set. Furthermore, to eliminate reliance on specific reward models, we conduct additional experiments using CLIPScore \cite{hessel2021clipscore}, a general-purpose explicit semantic evaluation reward model, as detailed in Sec.~\ref{sec:more_objective_evaluation}, where consistent performance gains are observed. For the WISE benchmark, training is constructed on the Knowledge-10K dataset, which encompasses extensive world knowledge and necessitates annotation by experts or flagship-class multimodal large language models (MLLMs). Although open-source MLLMs represent a viable alternative, we exclude NO-RL from this experiment to rigorously evaluate the scalability of the proposed method and its capability for large-scale knowledge acquisition without reward-based feedback. Consequently, the learning process relies solely on DK-DM. The experiments demonstrate the remarkable learning efficacy.

\section{Knowledge-10K}
\label{sec:B}

\noindent \textbf{Format.} Each entry in Knowledge-10K primarily consists of an \emph{implicit prompt} and its corresponding \emph{explicit prompt}. The former necessitates the retrieval and reasoning of deep-seated world knowledge, while the latter serves as its intuitive counterpart, which circumvents logical complexity by directly describing the visual content. Below are some representative data samples.

\begin{tcolorbox}[
    colback=MorandiBlue, 
    colframe=MorandiBlueFrame, 
    arc=4pt, 
    boxrule=1pt, 
    halign=flush left, 
    left=5pt, right=5pt, top=5pt, bottom=5pt, 
    % title=A,
]
\small{
\{\\
\qquad``prompt\_id'': 55,\\
\qquad``Impl\_prompt'': ``A monkey from China with a blue face and golden fur, adapted to cold climates'',\\
\qquad``Expl\_prompt'': ``Golden Snub-nosed Monkey.'',\\
\qquad``Category'': ``Cultural knowledge''\\
\},\\
\{\\
\qquad``prompt\_id'': 382,\\
\qquad``Impl\_prompt'': ``Lionel Messi's iconic moment'',\\
\qquad``Expl\_prompt'': ``Lionel Messi lifting the FIFA World Cup trophy with the Argentina national team.'',\\
\qquad``Category'': ``Cultural knowledge''\\
\},\\
}
\end{tcolorbox}

\noindent \textbf{Data review.} To prevent performance degradation caused by excessively long context, we provide Gemini 3 Pro with only 10 data entries at a time for review. The instruction used for data review during the construction of the Knowledge-10K dataset is provided as follows:

\begin{tcolorbox}[
    colback=MorandiBlue, 
    colframe=MorandiBlueFrame, 
    arc=4pt, 
    boxrule=1pt,
    halign=flush left, 
    left=5pt, right=5pt, top=5pt, bottom=5pt, 
    % title=A,
    enhanced jigsaw,
    breakable,
]
\small{
You are an expert prompt reviewer for text-to-image (T2I) generative models, possessing extensive world knowledge spanning from natural science to cultural commonsense.\\
\textbf{Input:}\\
You will receive a JSON array of data entries. Each entry contains:\\
- ``prompt\_id'': unique identifier.\\
- ``Impl\_prompt'': an implicit description that requires world knowledge to understand what specific thing it refers to.\\
- ``Expl\_prompt'': a rewritten explicit version of Impl\_prompt, it should be a direct, image-centric prompt that can be understood without any external knowledge, suitable for T2I generative models.\\
- ``Category'': the category of the entry.\\
\textbf{Review criteria:}\\
For each entry, check:\\
1. \textbf{Knowledge correctness}: Does the Expl\_prompt correctly correspond to the subject described in the Impl\_prompt? Please engage in thorough reflection and verification, leveraging your broad world knowledge to ensure accuracy.\\
2. \textbf{Image-centric}: Is the Expl\_prompt a direct, visual description suitable for T2I models (no abstract concepts, no explanations)?\\
3. \textbf{Completeness}: Does the Expl\_prompt capture the key visual elements implied by Impl\_prompt?\\
If an Expl\_prompt is basically reasonable, \textbf{keep it as is}, do not be overly strict. Only rewrite when there is a clear error or significant mismatch.\\
\textbf{Output:}\\
You MUST return a valid JSON object with exactly this structure:\\
\{\\
\qquad``prompt\_id'': 1,\\
\qquad``Expl\_promp'': ``Original or rewritten Expl\_prompt'',\\
\qquad``modified'': false,\\
\qquad``reason'': ``''\\
\},\\
\{\\
\qquad``prompt\_id'': 2,\\
\qquad``Expl\_prompt'': ``Rewritten Expl\_prompt here'',\\
\qquad``modified'': true,\\
\qquad``reason'': ``Brief explanation of why it is changed''\\
\}\\
\textbf{Rules:}\\
- You MUST return ALL entries, do NOT skip any entry.\\
- The number of items in results MUST equal the number of input entries.\\
- ``modified'' is true if you change the Expl\_prompt, false if you kept it as is.\\
- ``reason'' should be empty ``'' if not modified, a brief explanation if modified.\\
- Return ONLY the JSON object, no other text.
}
\end{tcolorbox}

\section{Additional experimental results}
\label{sec:C}

\subsection{User study}

We conduct a subjective user study involving 7 volunteers, all of whom hold at least a bachelor’s degree in engineering. Participants are asked to select the images that exhibit the highest scientific accuracy and semantic alignment, with 10 questions per participant. Participants are permitted to consult external reference materials for verification. A quantitative analysis of the voting results is presented in Table~\ref{tab:user-study}, which demonstrates that our method performs favorably against the other counterparts.

\begin{table}
\caption{\textbf{User study.} \textsc{EruDiff} performs over other counterparts.}
\label{tab:user-study}
\centering
\resizebox{0.525\linewidth}{!}{
\begin{tabular}{@{}lc@{}}
% \begin{tabular}{lcccccccc}
\toprule
\multicolumn{1}{c}{\bf Methods} & \makecell{\bf User study} \\
\midrule
FLUX.1-dev \cite{flux2024} & 1.43\% \\
% FLUX.1-dev + prompt rewrite (GPT-4o) & \\
FLUX.1-dev + SFT \cite{li2025science} & 1.43\% \\
FLUX.1-dev + OFT \cite{li2025science} & 7.14\% \\
FLUX.1-dev + GRPO \cite{liu2025flow} & 21.43\% \\
% \midrule
FLUX.1-dev + \textsc{EruDiff} (Ours) & \textbf{68.57\%} \\
\bottomrule
\end{tabular}
}
\end{table}

\subsection{More objective evaluation}
\label{sec:more_objective_evaluation}

\textbf{More results utilizing CLIPScore.} As demonstrated in Table~\ref{tab:more-science-t2i}, the integration of NO-RL further enhances the performance of the proposed method. Furthermore, the employment of CLIPScore, a generalized reward model for explicit semantic evaluation, yields measurable performance gains, thereby partially circumventing the necessity for specialized reward models.

\begin{table}
\caption{\textbf{More objective evaluation on \textsc{Science}-T2I.}}
\label{tab:more-science-t2i}
\centering
\resizebox{0.85\linewidth}{!}{
\begin{tabular}{@{}lc@{\quad}c@{}}
% \begin{tabular}{lcccccccc}
\toprule
\multicolumn{1}{c}{\bf Methods} & \makecell{\bf Science-T2I S\\ \textsc{SciScore}} & \makecell{\bf Science-T2I C\\ \textsc{SciScore}} \\
\midrule
FLUX.1-dev \cite{flux2024} & 23.31 & 27.54 \\
FLUX.1-dev + prompt rewrite (GPT-4o) & 32.90 & 33.19 \\
FLUX.1-dev + SFT \cite{li2025science} & 28.64 & 30.01 \\
FLUX.1-dev + OFT \cite{li2025science} & 30.95 & 32.31 \\
FLUX.1-dev + GRPO \cite{liu2025flow} & 31.68 & 32.19 \\
\midrule
FLUX.1-dev + \textsc{EruDiff} (CLIPScore) & {32.93} & {33.44} \\
FLUX.1-dev + \textsc{EruDiff} (\textsc{SciScore}) & {33.70} & {34.54} \\
\midrule
FLUX.1-dev + \textsc{EruDiff} (w/o NO-RL) & {31.99} & {32.82} \\
FLUX.1-dev + \textsc{EruDiff} (w/ NO-RL + positive) & {33.83} & {34.19} \\
\bottomrule
\end{tabular}
}
\end{table}

\noindent \textbf{More results on NO-RL.} As shown in Table~\ref{tab:more-science-t2i}, we provide additional ablation study regarding NO-RL, demonstrating that its incorporation further enhances performance. Furthermore, as noted in Sec. 3.2, the integration of redundant positive sample learning yields no further significant performance gains. In contrast to the filtering strategy for negative samples, which utilizes implicit prompts, positive samples are filtered through the use of explicit prompts.

\begin{table}
\caption{\textbf{More objective evaluation on WISE.}}
\label{tab:more-wise}
\centering
\resizebox{0.98\linewidth}{!}{
\begin{tabular}{@{}lccccccc@{}}
% \begin{tabular}{lcccccccc}
\toprule
\multicolumn{1}{c}{\bf Methods} & {Cultural} & {Time} & {Space} & {Biology} & {Physics} & {Chemistry} & \textbf{Overall} \\
\midrule
FLUX.1-dev & 0.48 & 0.58 & 0.62 & 0.42 & 0.51 & 0.35 & 0.50 \\
% \midrule
+ \textsc{EruDiff} (Knowledge-10K) & {0.66} & {0.65} & {0.72} & {0.54} & {0.64} & {0.49} & {0.64} \\
+ \textsc{EruDiff} (WISE) & {0.74} & {0.71} & {0.83} & {0.61} & {0.73} & {0.72} & {0.73} \\
\bottomrule
\end{tabular}}
\end{table}

\noindent \textbf{More quantitative results on WISE .} In Table 1 of the main paper, we provide an objective evaluation of \textsc{EruDiff}, trained on Knowledge-10K, using the WISE benchmark. Furthermore, as discussed in Sec.~\ref{sec:more_results_on_distribution_shift}, \textsc{EruDiff} is capable of incremental learning while preserving pre-trained knowledge. To explore the performance ceiling of \textsc{EruDiff}, and to account for the potential distribution gap between Knowledge-10K and WISE that might constrain evaluation metrics, we conduct training directly on the WISE benchmark. Specifically, we utilize evaluation prompts as implicit prompts and their corresponding GPT-4o-based rewrites as explicit prompts. The relevant results, presented in Table~\ref{tab:more-wise}, indicate that the performance of \textsc{EruDiff} is further unleashed.

\subsection{More results on distribution shift}
\label{sec:more_results_on_distribution_shift}

\begin{figure}
\centering
\setlength{\belowcaptionskip}{-0.4cm}
\includegraphics[width=0.85\linewidth]{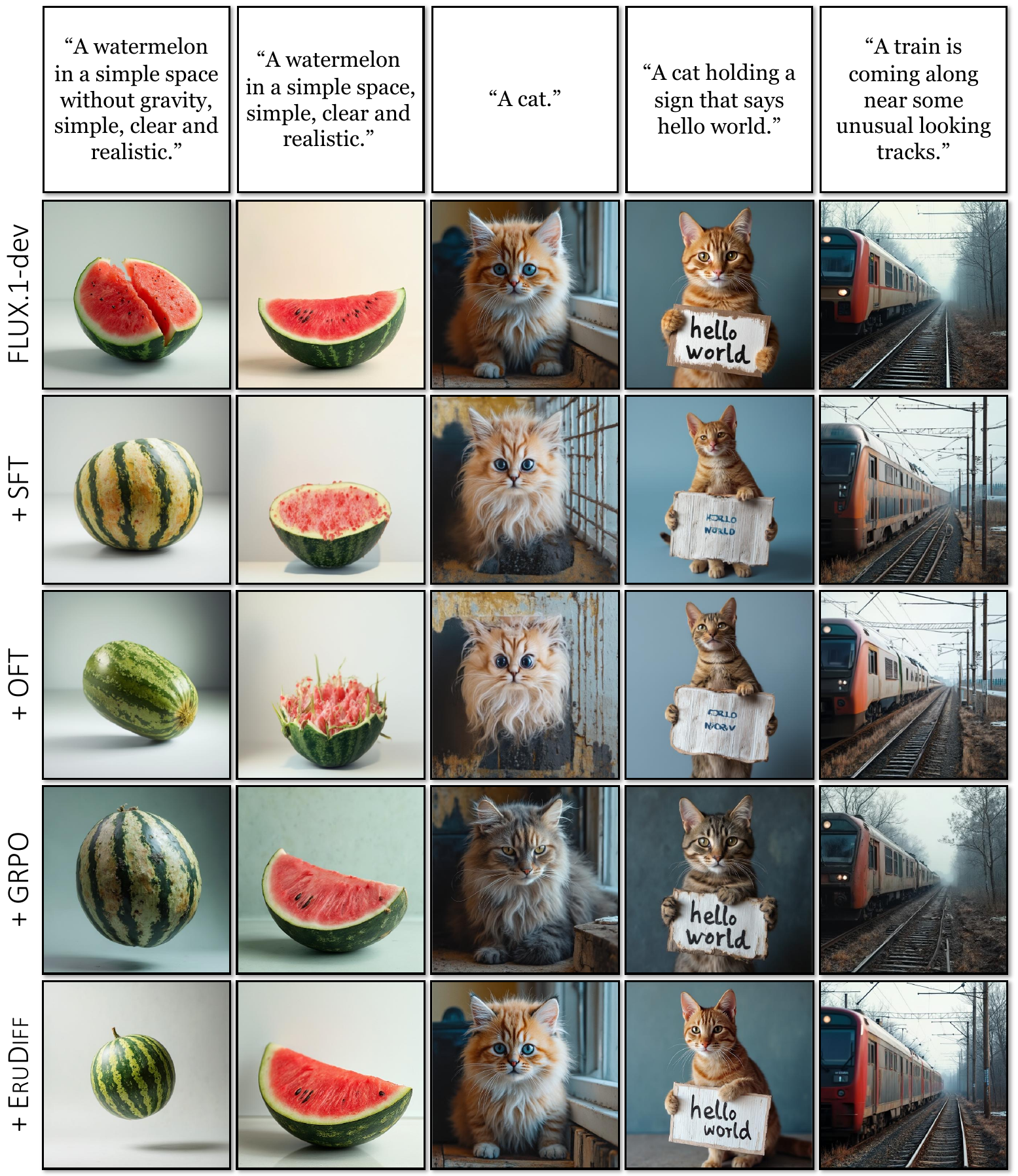}
\caption{\textbf{More results} on distribution shift. The proposed method effectively mitigates the risk of distribution shift, circumventing knowledge erosion and the degradation of color and texture.}
\label{fig:x_distribution_shift}
\end{figure}

The reinforcement learning fine-tuning process inevitably leads to distribution shift, which poses inherent risks of semantic degradation and the erosion of pre-trained knowledge. Our method effectively mitigates these limitations by leveraging a pure text corpus complemented by an anti-forgetting knowledge consolidation mechanism designed to prevent forgetting. As illustrated in Fig.~\ref{fig:x_distribution_shift}, SFT and RL-based methods frequently induce varying degrees of pre-trained knowledge erosion alongside color and texture degradation, exemplified by ``unrealistic watermelons'' and ``color deviations in cat''. Conversely, our approach effectively preserves foundational capabilities during the learning process, even for concepts absent from the training phase, such as ``cat'' and ``train''.

\subsection{Additional visual results}

\begin{figure}
\centering
\includegraphics[width=1.\linewidth]{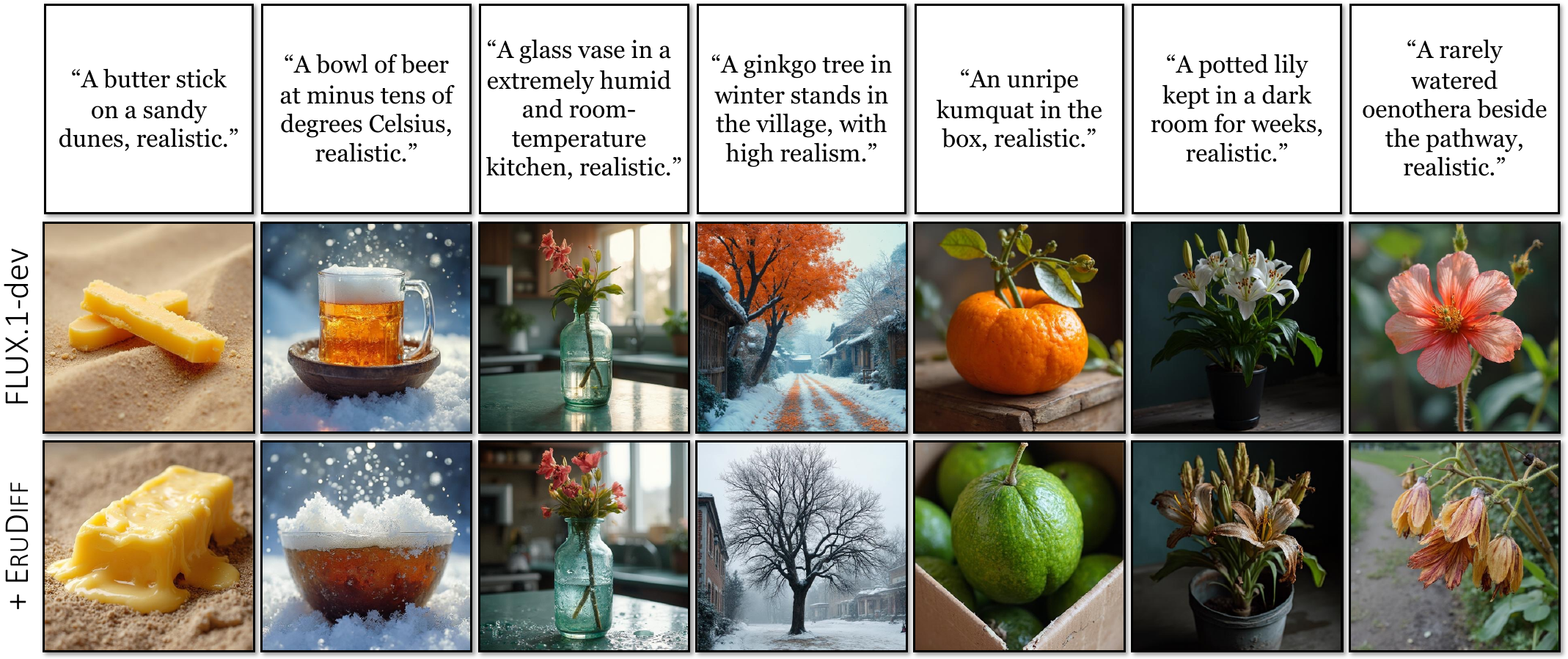}
\includegraphics[width=1.\linewidth]{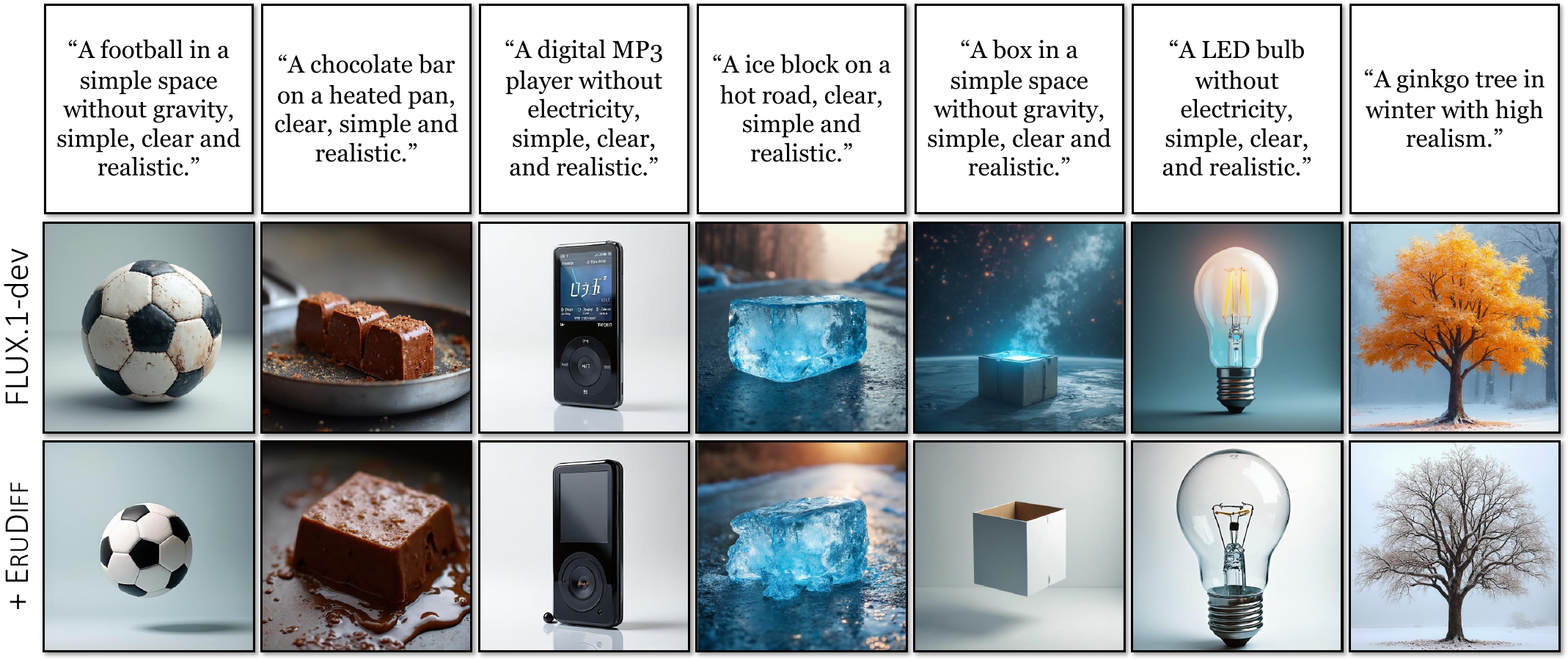}
\includegraphics[width=1.\linewidth]{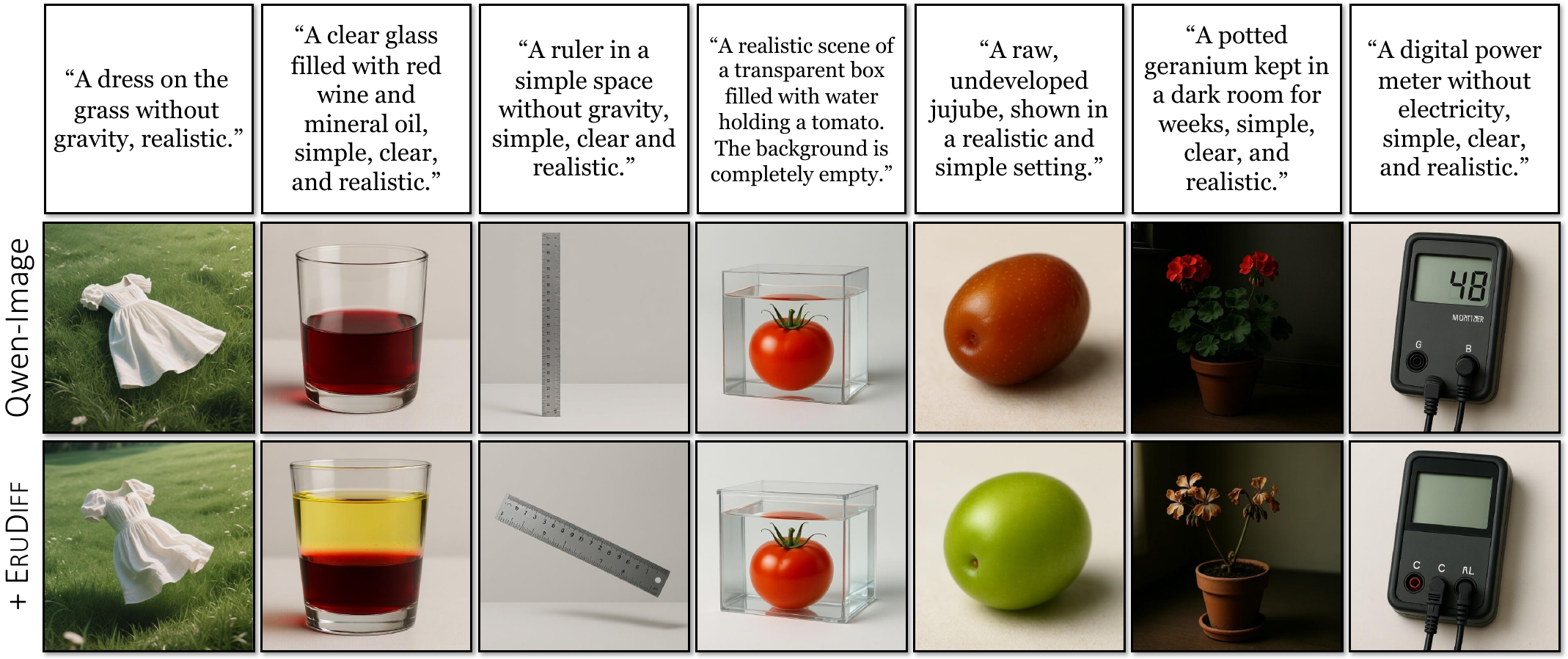}
\caption{\textbf{More visual results} synthesized by \textsc{EruDiff} on \textsc{Science}-T2I.}
\label{fig:x_more_visual_results}
\end{figure}

\begin{figure}
\centering
\includegraphics[width=1.\linewidth]{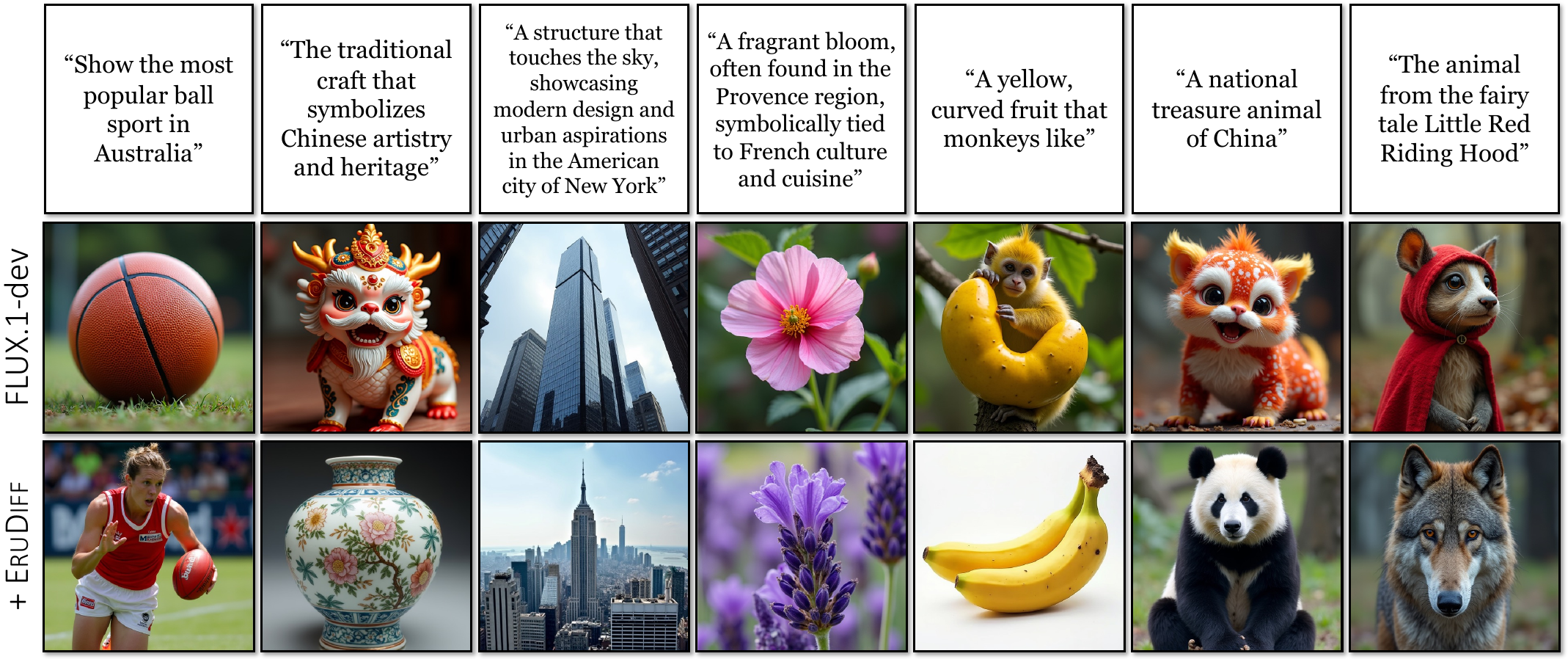}
\includegraphics[width=1.\linewidth]{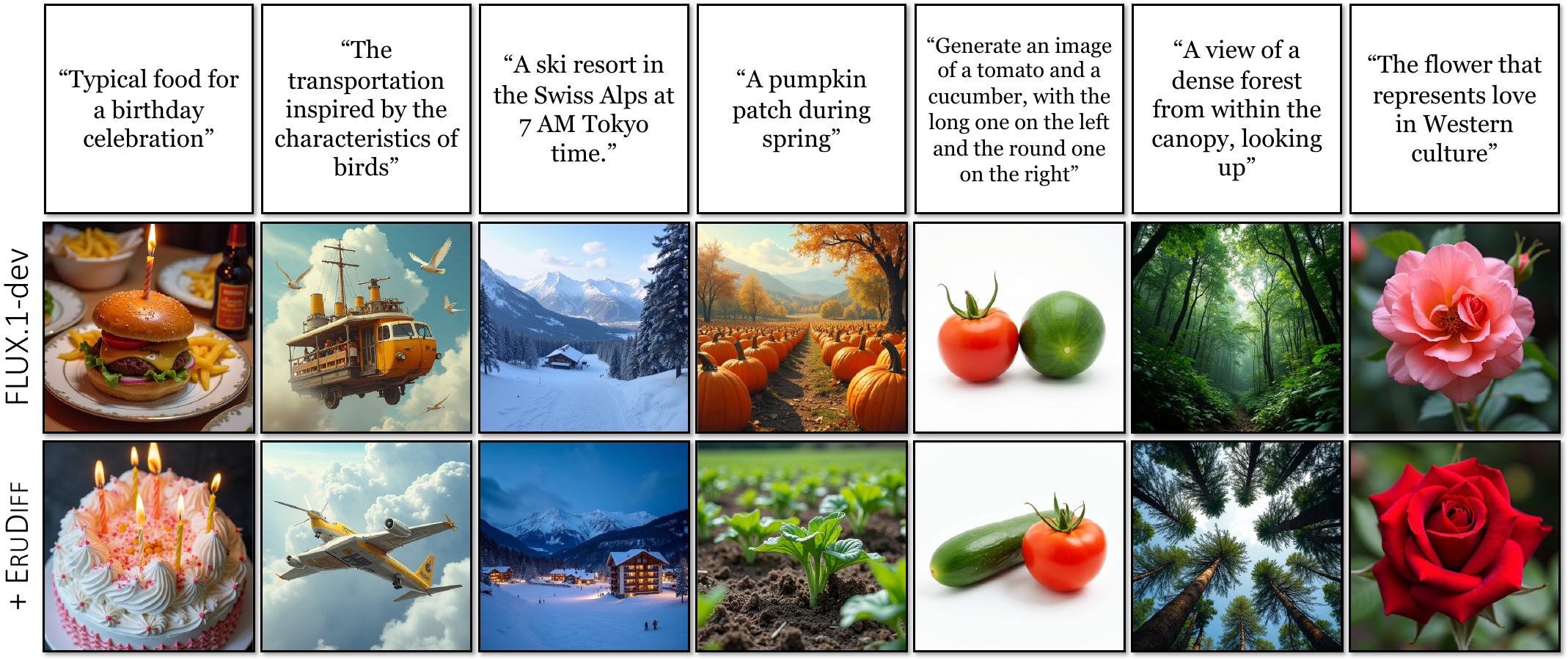}
\includegraphics[width=1.\linewidth]{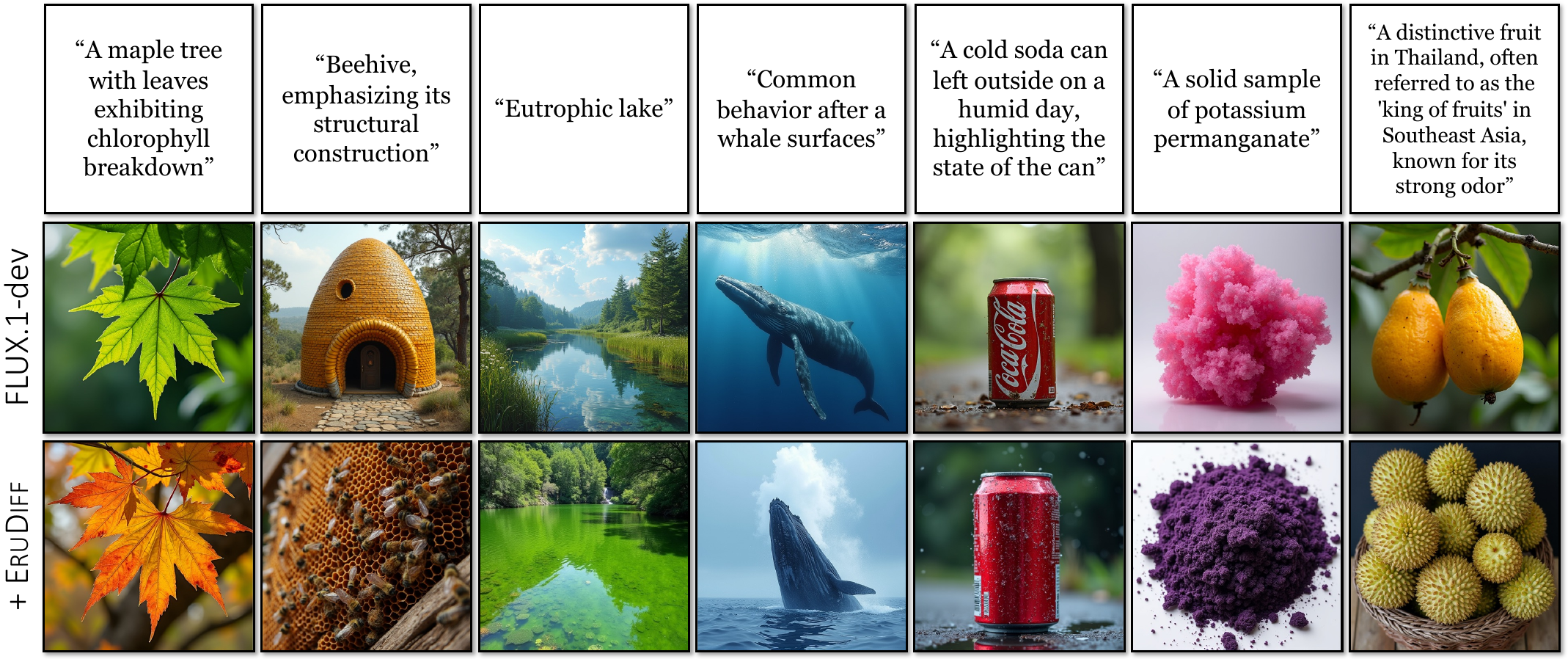}
\caption{\textbf{More visual results} synthesized by \textsc{EruDiff} on WISE.}
\label{fig:x_wise}
\end{figure}

Fig.~\ref{fig:x_more_visual_results} and \ref{fig:x_wise} show additional visual results of our approach achieved on on the Science-T2I and WISE benchmarks, respectively.

\end{document}